\documentclass[10pt,twocolumn,letterpaper]{article}

\usepackage[pagenumbers]{cvpr} 

\definecolor{cvprblue}{rgb}{0.21,0.49,0.74}
\usepackage[pagebackref,breaklinks,colorlinks,allcolors=cvprblue]{hyperref}
\usepackage{multirow}
\usepackage{float}
\usepackage{placeins}   
\usepackage{stfloats}   
\usepackage{tikz}
\usetikzlibrary{positioning}

\usepackage[font=small,skip=3pt]{caption}
\usepackage{subcaption}
\usepackage{booktabs}
\usepackage{microtype}
\usepackage{float}
\usepackage{placeins}
\usepackage{afterpage} 

\def\paperID{*****} 
\def\confName{CVPR}
\def\confYear{2026}

\title{RePose-NeRF: Robust Radiance Fields for Mesh Reconstruction under Noisy Camera Poses}

\author{Sriram Srinivasan\\
Bellatrix Aerospace\\
{\tt\small sriram@bellatrix.aero}
\and
Gautam Ramachandra \\
Bellatrix Aerospace\\
{\tt\small gautam@bellatrix.aero}
}

\begin{document}
\maketitle
\begin{abstract}
Accurate 3D reconstruction from multi-view images is essential for downstream robotic tasks such as navigation, manipulation, and environment understanding. However, obtaining precise camera poses in real-world settings remains challenging, even when the calibration parameters are known. This limits the practicality of existing NeRF-based methods that rely heavily on accurate extrinsic estimates. Furthermore, their implicit volumetric representations differ significantly from the widely adopted polygonal meshes, making rendering and manipulation inefficient in standard 3D software. In this work, we propose a robust framework that reconstructs high-quality, editable 3D meshes directly from multi-view images with noisy extrinsic parameters. Our approach jointly refines camera poses while learning an implicit scene representation that captures fine geometric detail and photorealistic appearance. The resulting meshes are compatible with common 3D graphics and robotics tools, enabling efficient downstream use. Experiments on standard benchmarks demonstrate that our method achieves accurate and robust 3D reconstruction under pose uncertainty, bridging the gap between neural implicit representations and practical robotic applications.
\end{abstract}

\section{Introduction}

Reconstructing accurate 3D models from multi-view images is a fundamental problem in computer vision, with wide-ranging applications in robotics, augmented reality, and digital simulation. In robotics, high-fidelity 3D models play a crucial role in perception, navigation, and manipulation, enabling autonomous systems to reason about object geometry and spatial relationships. Traditional 3D reconstruction techniques based on multi-view geometry and photogrammetry~\cite{7780814, yao2018mvsnetdepthinferenceunstructured} require dense viewpoints and precise camera calibration, often limiting their scalability and robustness under real-world conditions.

Recent advances in neural implicit representations~\cite{Park_2019_CVPR, mescheder2019occupancynetworkslearning3d} have transformed 3D scene reconstruction, with Neural Radiance Fields (NeRF)~\cite{mildenhall2020nerf, barron2021mipnerf, Chen2022ECCV, mueller2022instant} emerging as a powerful framework for modeling complex scenes from 2D observations. NeRF represents a scene as a continuous volumetric function that maps a 3D location and viewing direction to emitted color and volume density. By optimizing this function through differentiable volume rendering, NeRF learns a neural implicit representation capable of synthesizing photorealistic novel views. However, this process is critically dependent on accurate camera poses to maintain geometric consistency across viewpoints. Inaccurate or noisy poses lead to misaligned radiance fields, causing artifacts such as ghosting, texture drift, and geometric distortions.

Obtaining precise camera poses in practice remains highly challenging. Pose estimates derived from Structure-from-Motion (SfM)~\cite{schoenberger2016mvs, schoenberger2016sfm} or SLAM pipelines often degrade in the presence of low-texture regions, motion blur, or varying illumination. Consequently, NeRF-based methods that assume perfect extrinsic parameters struggle to generalize to real-world data. Pose-robust variants such as BARF~\cite{lin2021barf} and SPARF~\cite{sparf2023} attempt to jointly refine camera poses and radiance fields during training, but they remain computationally expensive and operate solely in the implicit volumetric domain, limiting their practical usability.

Moreover, the implicit representation used by NeRF is incompatible with the polygonal mesh formats required in most downstream robotics and graphics applications. Polygonal meshes are the standard for physics simulation, rendering, and interaction due to their efficiency and interoperability with existing 3D tools. Extracting high-quality meshes from implicit NeRF fields typically involves post-processing steps such as Marching Cubes~\cite{lorensen1987marchingcubes}, SDF-based differentiable surface extraction~\cite{wang2021neus, yariv2021volume}, or mesh-regularized reconstruction methods~\cite{li2023neuralangelo, neuralwarp}. However, these methods become unreliable when the underlying density field is misaligned by pose noise, leading to artifacts and inconsistent geometry.

To address these challenges, we propose RePose-NeRF (Robust Radiance Fields for Mesh Reconstruction under Noisy Camera Poses), a robust 3D reconstruction framework that generates high-quality and editable 3D meshes directly from multi-view images with noisy extrinsic parameters. RePose-NeRF jointly refines camera poses through differentiable bundle adjustment while learning an implicit neural representation that captures accurate geometry and photorealistic appearance. Built upon the Instant-NGP~\cite{mueller2022instant} architecture, our method enables efficient gradient propagation through camera parameters, achieving rapid convergence and stable optimization even under pose uncertainty. Once trained, the learned implicit field is converted into a polygonal mesh via surface extraction, producing view-consistent, high-fidelity 3D models compatible with standard graphics and robotics software. This integration allows seamless deployment of reconstructed assets in downstream perception, manipulation, and simulation pipelines.

\noindent \textbf{Contributions.} The key contributions of this work are summarized as follows:
\begin{itemize}
    \item We propose a pose-robust 3D reconstruction framework that learns accurate scene geometry directly from multi-view images with noisy extrinsic parameters.
    \item We introduce a joint optimization strategy that refines camera poses and radiance field parameters during training, improving geometric consistency and reconstruction stability.
    \item We enable efficient and editable 3D reconstruction by extracting high-fidelity polygonal meshes that preserve photorealistic detail and are compatible with standard 3D graphics and robotics simulation tools.
\end{itemize}

\section{Related Work}

\subsection{Scene Representation using NeRF}

Neural Radiance Fields (NeRF)~\cite{mildenhall2020nerf} and its numerous variants have demonstrated remarkable capability in representing complex 3D scenes from 2D images, achieving high-quality novel view synthesis with fine geometric and appearance details. However, despite their superior rendering performance, classical NeRF models face several challenges in real-world applications. Similar to Structure-from-Motion (SfM) pipelines, NeRF requires accurate camera poses to learn a consistent implicit scene representation. In practice, camera poses are often estimated using SfM tools such as COLMAP~\cite{schoenberger2016sfm, schoenberger2016mvs}, which are susceptible to errors in scenes with limited parallax, low texture, or inconsistent illumination, often resulting in incomplete or distorted reconstructions.

To overcome this limitation, several approaches jointly optimize camera poses and NeRF parameters during training. Early works such as NeRF--~\cite{wang2022nerfneuralradiancefields} and Self-Calibrating NeRF~\cite{SCNeRF2021} introduced pose refinement modules that learn extrinsic parameters alongside scene reconstruction. Bundle-Adjusting Radiance Fields (BARF)~\cite{lin2021barf} further proposed a coarse-to-fine optimization strategy that refines both pose and radiance field parameters, leading to more stable convergence. While effective, BARF assumes smooth camera trajectories and struggles under abrupt viewpoint changes. Gaussian-Activated Radiance Fields (GARF)~\cite{chng2022gaussian} stabilized training by replacing sinusoidal encodings with Gaussian activations, improving pose refinement but at high computational cost for large-scale scenes. SPARF [26] (Neural Radiance Fields from Sparse and Noisy Poses) tackles the problem of training NeRFs with limited and inaccurate camera poses by enforcing multi-view consistency constraints across different viewpoints. While this improves robustness under sparse and noisy pose conditions, it also increases the overall optimization complexity during training. These methods collectively focus on enhancing NeRF’s robustness to pose noise and initialization errors.

While these methods improve robustness to noisy poses, they remain computationally intensive due to iterative pose updates and volumetric rendering. To accelerate training, subsequent works explored alternative scene representations. Direct Voxel Grid Optimization (DVGO)~\cite{SunSC22} replaced the MLP with learnable voxel grids, enabling faster reconstruction but sacrificing fine-detail fidelity and requiring large memory. Instant-NGP~\cite{mueller2022instant} achieved substantial efficiency gains by introducing a multi-resolution hash-grid encoding that compactly stores spatial features across scales, allowing near real-time training and rendering.

Despite these advancements, existing NeRF-based methods still rely on accurate or refined poses and produce implicit volumetric representations that are not directly compatible with standard 3D modeling tools, limiting their applicability in real-world robotic and simulation settings.

\subsection{Surface Mesh Reconstruction and Extraction}

Extracting explicit 3D surfaces from implicit neural representations remains a challenging task due to NeRF’s volumetric rendering formulation. Traditional approaches often rely on template meshes or fixed topologies, limiting flexibility for complex scenes.  
Recent works have focused on deriving surface geometry directly from radiance fields or signed distance representations.  
NeuS~\cite{wang2021neus} and VolSDF~\cite{yariv2021volume} reinterpret volumetric rendering through signed distance functions (SDFs), enabling smooth surface extraction via differentiable rendering. Nvdiffrec~\cite{Munkberg_2022_CVPR} combines differentiable rasterization and marching tetrahedrons for gradient-based optimization of mesh surfaces, allowing fine-grained geometric reconstruction. Other approaches, such as NeuralWarp~\cite{neuralwarp} and Neuralangelo~\cite{li2023neuralangelo}, integrate neural feature fields with differentiable surface extraction pipelines to recover high-fidelity meshes with fine details. Classical techniques such as Marching Cubes \cite{lorensen1987marchingcubes} remain widely used for extracting polygonal meshes from implicit fields due to their simplicity and efficiency; however, when applied to NeRFs, they often produce coarse surface structures that require additional refinement because of sensitivity to density thresholds and limited geometric precision.

Despite these advances, achieving consistent mesh reconstruction from radiance fields remains computationally demanding, particularly for large-scale or unbounded scenes. Our method aims to bridge this gap by integrating efficient radiance field learning with lightweight surface extraction strategies.

\section{Preliminaries}

In this section, we introduce the notations and fundamental formulations of Neural Radiance Fields (NeRF)~\cite{mildenhall2020nerf}, along with the multi-resolution hash-grid encoding~\cite{mueller2022instant} employed in our framework.

\subsection{Camera Pose and Parameters}
Let $\mathbf{P}_i = [\mathbf{R}_i \,|\, \mathbf{t}_i] \in \text{SE}(3)$ denote the camera-to-world transformation matrix for the $i^{\text{th}}$ view, where $\mathbf{R}_i \in \text{SO}(3)$ and $\mathbf{t}_i \in \mathbb{R}^3$ represent rotation and translation components, respectively.  
The learnable camera pose parameters are denoted as $\boldsymbol{\theta}_i \in \mathfrak{se}(3)$, consisting of three parameters for rotation and three for translation.  
This Lie algebra representation allows smooth optimization of 6-DoF camera transformations during training.

\subsection{Scene Representation via NeRF}
Neural Radiance Fields (NeRF)~\cite{mildenhall2020nerf} model a continuous volumetric scene as a function that maps a 3D spatial coordinate $\mathbf{x} \in \mathbb{R}^3$ and a viewing direction $\mathbf{v} \in \mathbb{S}^2$ to an RGB color $\mathbf{c} \in [0,1]^3$ and a volume density $\sigma \in \mathbb{R}_{\ge 0}$.  
This mapping is parameterized by a Multi-Layer Perceptron (MLP):
\begin{equation}
[\mathbf{c}, \sigma] = \text{MLP}\big(\gamma_x(\mathbf{x}), \gamma_v(\mathbf{v})\big),
\label{eq:nerf_mlp}
\end{equation}
where $\gamma_x(\cdot)$ and $\gamma_v(\cdot)$ denote the positional and directional encoding functions, respectively.

\subsection{Volumetric Rendering}
Given a camera ray $\mathbf{r}(t) = \mathbf{o} + t\mathbf{v}$, where $\mathbf{o}$ is the camera origin and $\mathbf{v}$ is the normalized direction, the pixel color is synthesized by integrating the radiance along the ray.  
For a neural field $f_\theta(\mathbf{x}, \mathbf{v})$, the expected color $\hat{\mathbf{C}}(\mathbf{r})$ is computed as:
\begin{equation}
\hat{\mathbf{C}}(\mathbf{r}) = \sum_{i=1}^{N} T_i \alpha_i \mathbf{c}_i,
\label{eq:rendering_color}
\end{equation}
where the transmittance $T_i$ and opacity $\alpha_i$ for the $i^{\text{th}}$ sample are defined as:
\begin{equation}
\alpha_i = 1 - \exp(-\sigma_i \delta_i), \quad
T_i = \exp\left(-\sum_{j=1}^{i-1} \sigma_j \delta_j\right),
\label{eq:rendering_weights}
\end{equation}
and $\delta_i = t_{i+1} - t_i$ represents the interval between consecutive depth samples.  
Eqs.~\ref{eq:rendering_color}--\ref{eq:rendering_weights} model light absorption and emission along each ray, enabling photorealistic view synthesis of the underlying scene geometry.

\subsection{Multi-Resolution Hash Encoding}
To accelerate the learning of radiance fields, we employ the multi-resolution hash-grid encoding~\cite{mueller2022instant}, which provides a compact yet expressive representation of 3D space.  
Each spatial coordinate $\mathbf{x} \in [0,1]^d$ is represented across $L$ grids of exponentially increasing resolution, forming a concatenated multi-scale feature vector.  
At level $l$, the grid resolution $N_l$ is defined as:

\begin{equation} N_l = \left\lfloor N_{\text{min}} \cdot b^l \right\rfloor, \quad \text{where} \quad b = \exp\left( \frac{\ln N_{\text{max}} - \ln N_{\text{min}}}{L - 1} \right). \label{eq:grid_resolution} 
\end{equation}

For each level, a scaled coordinate $\mathbf{x}_l = \mathbf{x} \cdot N_l$ determines the enclosing grid cell via its integer floor and ceil indices.  
Trilinear interpolation (for $d=3$) is applied over the surrounding $2^d$ vertices, with interpolation weights computed as:
\begin{equation}
w_l^i = x_l^i - \lfloor x_l^i \rfloor.
\label{eq:trilinear_weights}
\end{equation}

To maintain a fixed memory footprint, each grid level stores a feature table of size $\mathbf{T}$, independent of resolution.  
When the number of vertices exceeds $\mathbf{T}$, a spatial hash function (Eq.~\ref{eq:hash_function}) maps grid vertices to feature indices:
\begin{equation}
h(\mathbf{x}) = \left( \bigoplus_{i=1}^{d} (x_i \cdot \pi_i) \right) \bmod T,
\label{eq:hash_function}
\end{equation}
where $\mathbf{x} \in \mathbb{Z}^d$ is the integer vertex coordinate, $\bigoplus$ denotes the bitwise XOR operation, and $\pi_i$ are large primes (e.g., $\pi_1 = 1$, $\pi_2 = 2654435761$, $\pi_3 = 805459861$ for 3D).  
This hash mapping (Eqs.~\ref{eq:grid_resolution}--\ref{eq:hash_function}) introduces controlled collisions while preserving locality, yielding compact yet high-fidelity spatial features for efficient optimization.

\section{Proposed Method}
\label{sec:method}

We introduce a two-stage framework for reconstructing high-quality textured 3D surface meshes from multi-view images with noisy extrinsic parameters.  
Our design is conceptually inspired by the staged optimization paradigm proposed in NeRF2Mesh~\cite{tang2022nerf2mesh}, which separates neural field learning and mesh refinement to improve reconstruction stability and mesh quality.  
While both methods employ grid-based neural representations for efficiency, our framework differs fundamentally in its objectives and optimization strategy, as we explicitly integrate differentiable pose refinement in the first stage and extend the grid-based NeRF formulation \cite{mueller2022instant} to jointly optimize geometry, appearance, and camera parameters under pose uncertainty. Stage~1 performs coarse-to-fine joint pose and scene optimization, while Stage~2 refines the extracted mesh and texture to yield a high-fidelity, view-consistent, and editable 3D model.

\subsection{Stage 1: Efficient NeRF Learning and Pose Refinement}

We represent the scene using two continuous functions:  
a geometry function $f: \mathbb{R}^3 \rightarrow \mathbb{R}$ that maps a 3D position $\mathbf{x} \in \mathbb{R}^3$ to its signed distance, and an appearance function $c: \mathbb{R}^3 \times \mathbb{S}^2 \rightarrow [0,1]^3$ that maps a spatial point and viewing direction $\mathbf{v} \in \mathbb{S}^2$ to an RGB color.  
Both are parameterized by Multi-Layer Perceptrons (MLPs), consistent with the NeRF formulation in Eq.~\ref{eq:nerf_mlp}.

During this stage, we jointly optimize the neural scene representation and the noisy camera poses using a coarse-to-fine positional encoding strategy inspired by BARF~\cite{lin2021barf}.  
Following recent grid-based neural field methods~\cite{mueller2022instant}, we employ a multi-resolution hash-grid encoding (Eqs.~\ref{eq:grid_resolution}--\ref{eq:hash_function}) to embed spatial information efficiently across multiple scales.  
This enables accurate geometry and appearance learning while maintaining real-time training speed.  
The outputs of Stage~1 include refined camera poses, a coarse implicit field, and an initial mesh used for further refinement.

\subsubsection{Geometry Learning}

Instead of predicting raw volumetric density as in traditional NeRF~\cite{mildenhall2020nerf}, we represent scene geometry using a Signed Distance Function (SDF) following NeuS~\cite{wang2021neus} and VolSDF~\cite{yariv2021volume}.  
The zero-level set of the SDF defines the scene surface, allowing precise mesh extraction via Marching Cubes~\cite{lorensen1987marchingcubes}.  
The SDF is modeled using a shallow MLP over a multi-resolution hash encoding $E^{\text{geo}}(\cdot)$:
\begin{equation}
f(\mathbf{x}) = \text{MLP}\big(E^{\text{geo}}(\mathbf{x})\big),
\label{eq:sdf_mlp}
\end{equation}
where $f(\mathbf{x})$ denotes the signed distance at point $\mathbf{x}$.  
This grid-based encoding accelerates convergence and allows the network to capture fine geometric details efficiently.

To connect the SDF representation with volumetric rendering (Eqs.~\ref{eq:rendering_color}--\ref{eq:rendering_weights}), we adopt a sigmoid-based conversion to map SDF values to opacity:
\begin{equation}
\alpha = 
\frac{\psi(s \cdot \text{SDF}_{\text{prev}}) - \psi(s \cdot \text{SDF}_{\text{next}})}
{\psi(s \cdot \text{SDF}_{\text{prev}}) + \varepsilon},
\label{eq:sdf_opacity}
\end{equation}
where $\psi(\cdot)$ is the sigmoid function, $s = \exp(\beta)$ is a learnable sharpness parameter, and $\varepsilon$ ensures numerical stability.  
This formulation allows smooth gradients through the rendering process while maintaining sharp surface boundaries.

\subsubsection{Appearance Learning}

\begin{figure}[!t]
    \centering
    \begin{tabular}{ccc}
        \includegraphics[width=0.15\textwidth]{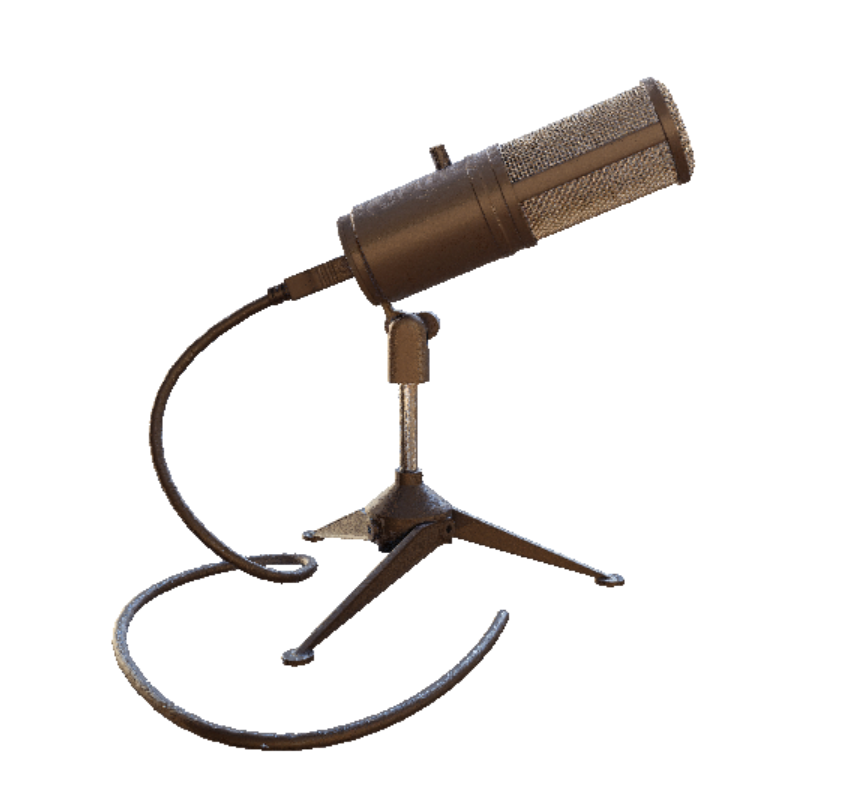} &
        \includegraphics[width=0.15\textwidth]{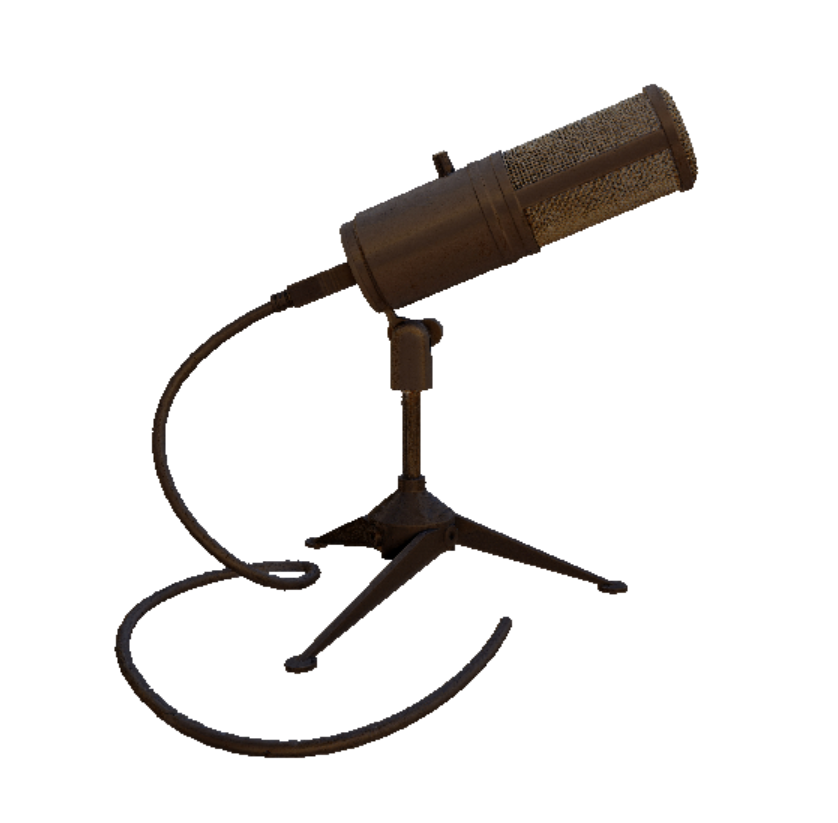} &
        \includegraphics[width=0.15\textwidth]{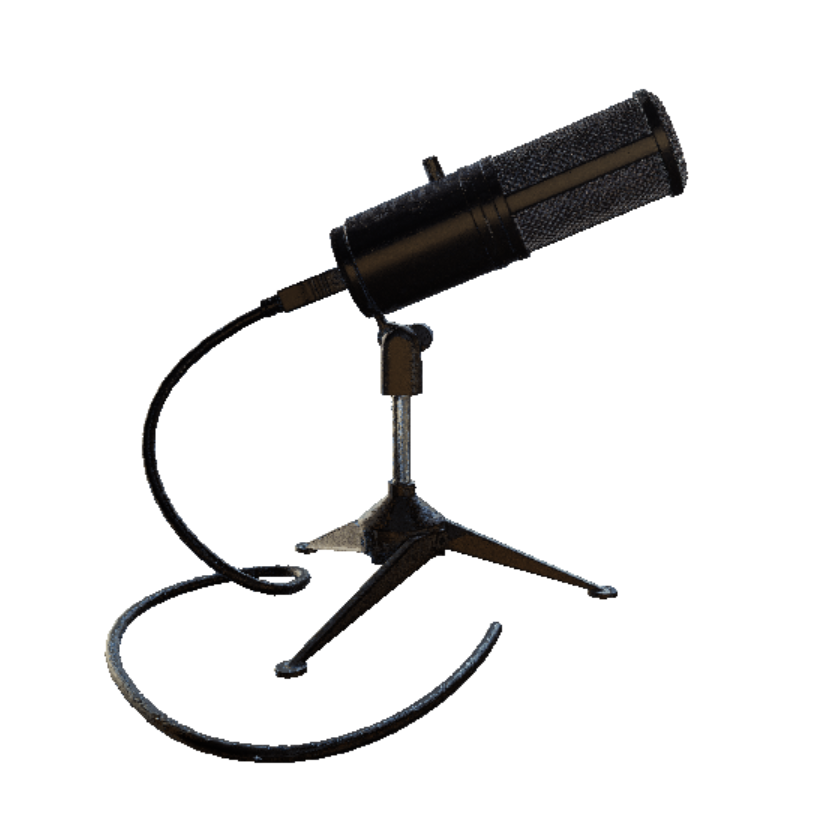} \\
        \footnotesize{Final Color} & \footnotesize{Diffuse} & \footnotesize{Specular} \\
    \end{tabular}
    \caption{Decomposition of the rendered image into diffuse and specular components. The final color image combines both components to produce the complete appearance.}
    \label{fig:render_decomposition}
\end{figure}

We model view-dependent appearance by decomposing the radiance field into diffuse and specular components, similar to Ref-NeRF~\cite{verbin2022refnerf} and NeRS~\cite{zhang2021ners}.  
Color is computed as:
\begin{align}
\mathbf{c}_d, \mathbf{f}_s &= \psi\!\left(\text{MLP}_{\text{diffuse}}\big(E^{\text{app}}(\mathbf{x})\big)\right), \label{eq:diffuse}\\
\mathbf{c}_s &= \psi\!\left(\text{MLP}_{\text{specular}}\big(\mathbf{f}_s, \mathbf{v}\big)\right), \label{eq:specular}\\
\mathbf{c} &= \mathbf{c}_d + \mathbf{c}_s, \label{eq:total_color}
\end{align}
where $\psi(\cdot)$ denotes the sigmoid activation, $\mathbf{f}_s$ is an intermediate specular feature vector, and $\mathbf{v}$ is the viewing direction.  
This decomposition facilitates explicit control over texture and lighting:  
(1) the diffuse term $\mathbf{c}_d$ can be directly converted into texture maps,  
(2) the specular term $\mathbf{c}_s$ supports realistic relighting, and  
(3) baked lighting textures can be reused in simulation and robotics applications. Figure~\ref{fig:render_decomposition} illustrates the individual diffuse and specular components, demonstrating how their combination reconstructs the complete appearance of the scene.

\subsubsection{Pose Refinement}

\begin{figure}[!t]
    \centering
    \begin{tabular}{ccc}
    \multicolumn{1}{c}{\shortstack{\footnotesize \textbf{Initial pose} \\ \footnotesize (0 iterations)}} &
    \multicolumn{1}{c}{\shortstack{\footnotesize \textbf{Intermediate pose} \\ \footnotesize (10,000 iterations)}} &
    \multicolumn{1}{c}{\shortstack{\footnotesize \textbf{Optimized pose} \\ \footnotesize (30,000 iterations)}} \\

        \\[-0.8em] 
        \includegraphics[width=0.125\textwidth]{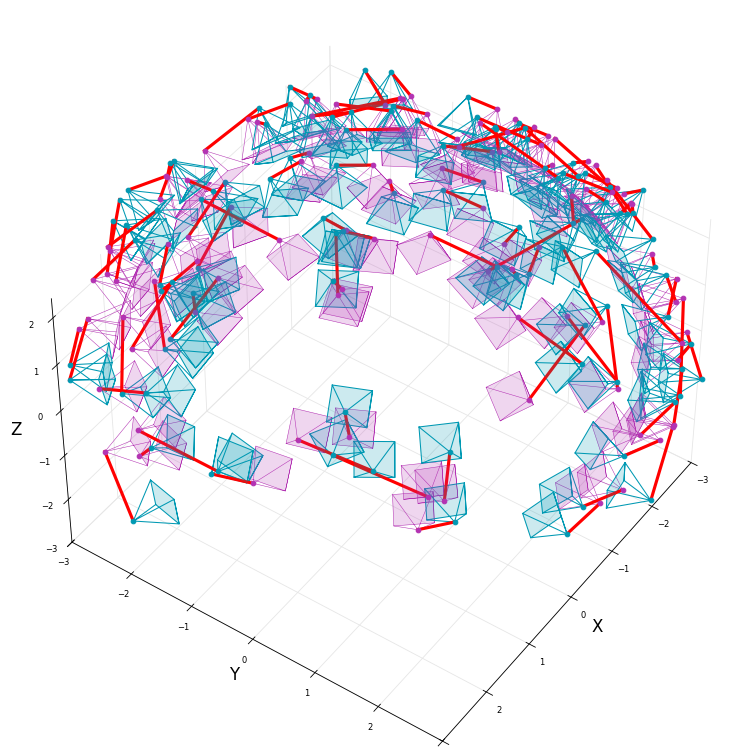} &
        \includegraphics[width=0.125\textwidth]{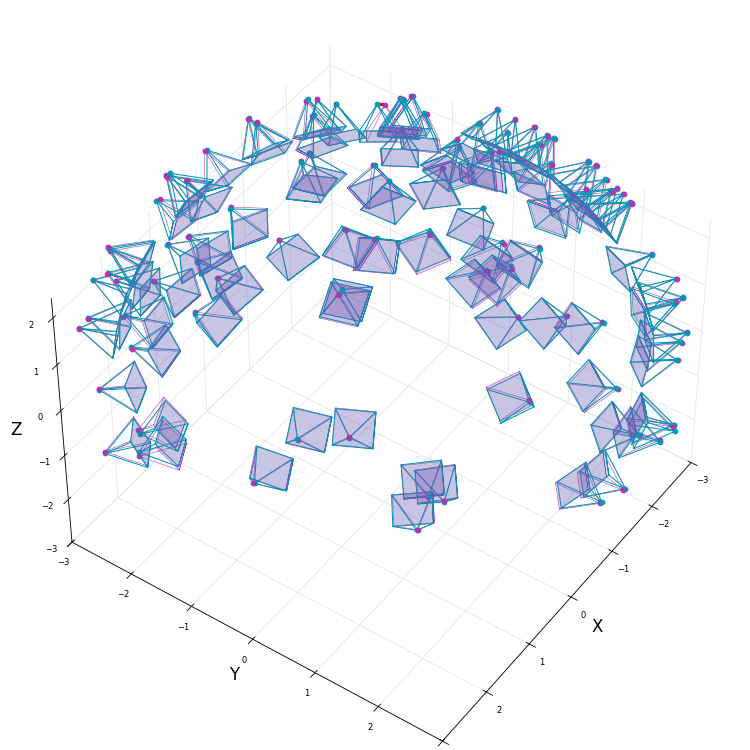} &
        \includegraphics[width=0.125\textwidth]{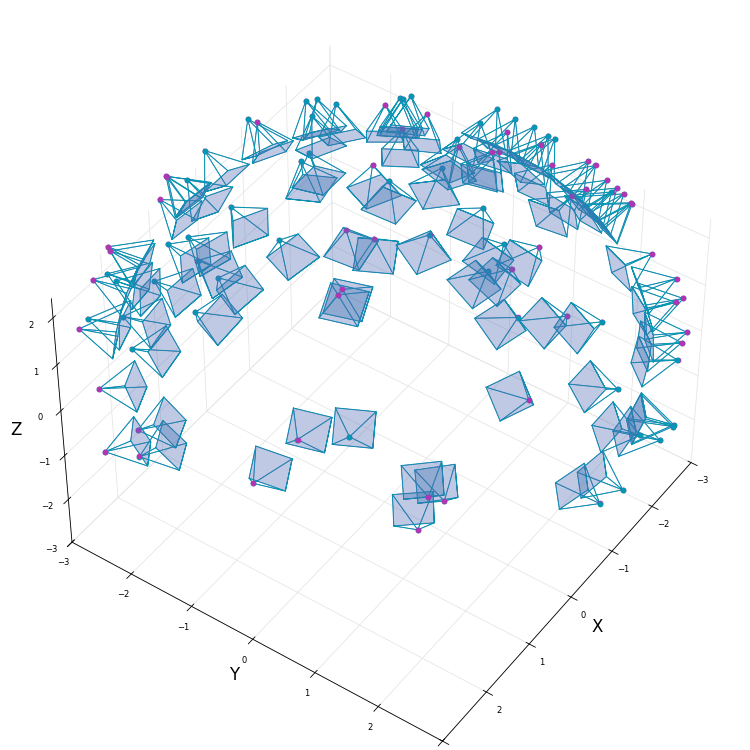} \\
        \includegraphics[width=0.125\textwidth]{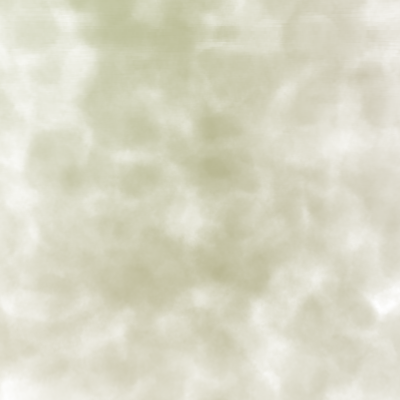} &
        \includegraphics[width=0.125\textwidth]{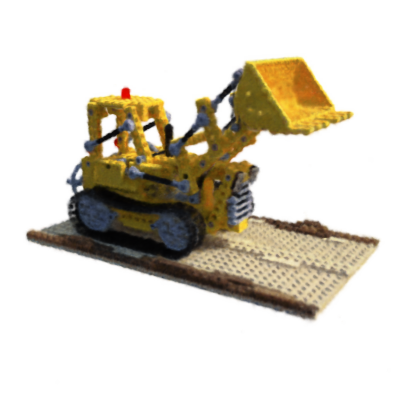} &
        \includegraphics[width=0.125\textwidth]{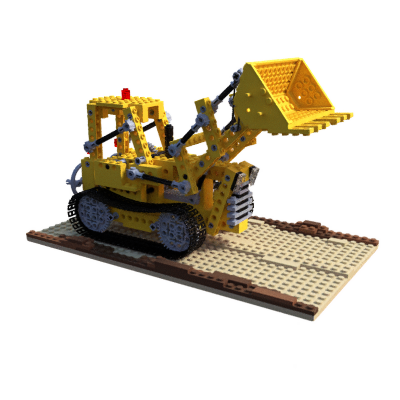} \\
        \includegraphics[width=0.125\textwidth]{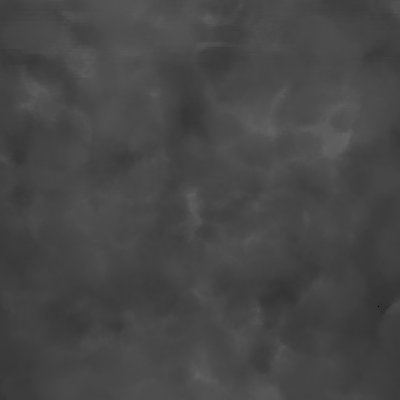} &
        \includegraphics[width=0.125\textwidth]{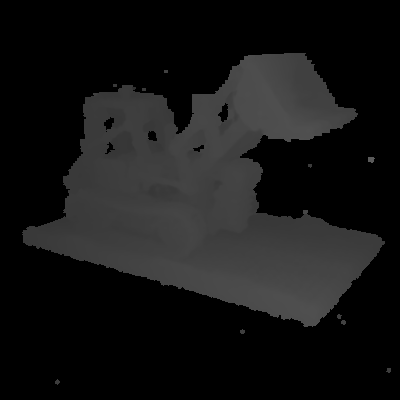} &
        \includegraphics[width=0.125\textwidth]{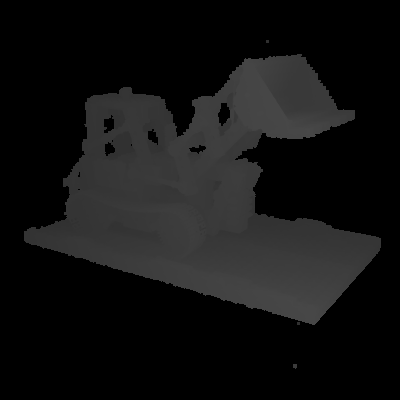} \\
    \end{tabular}
    \caption{Pose refinement results over optimization iterations. The initial estimates exhibit significant noise, which is progressively reduced as the optimization proceeds, leading to accurate and stable pose recovery after 30,000 iterations.}

    \label{fig:pose_refinement}
\end{figure}

Each camera pose is represented as an element of $\mathrm{SE}(3)$ and parameterized by a 6D vector $\boldsymbol{\xi}_i \in \mathfrak{se}(3)$ encoding rotation and translation.  
For each camera $i$, the refined pose is computed as:
\begin{equation}
\mathbf{T}_i = \exp(\hat{\boldsymbol{\xi}}_i), \quad
\mathbf{P}_i^{\text{refined}} = \mathbf{T}_i \cdot \mathbf{P}_i,
\label{eq:pose_refinement}
\end{equation}
where $\hat{\boldsymbol{\xi}}_i$ is the skew-symmetric matrix form of $\boldsymbol{\xi}_i$.  
 We apply a coarse-to-fine masking strategy over the hash-grid encoding levels, gradually introducing higher-resolution levels to refine finer pose components. 
This ensures smooth convergence and stability, especially under large initial pose noise, consistent with BARF~\cite{lin2021barf}. Figure \ref{fig:pose_refinement}. Pose refinement over optimization iterations. Initial noisy estimates are gradually corrected, achieving accurate and stable poses after 30,000 iterations.

\subsubsection{Loss Functions}

We jointly optimize geometry, appearance, and pose parameters using a weighted combination of losses:

\paragraph{Photometric Reconstruction Loss.}
\begin{equation}
\mathcal{L}_{\text{photo}} = 
\frac{1}{N} \sum_{i=1}^{N} 
\big\| \hat{\mathbf{C}}_i - \mathbf{c}_i \big\|_2^2,
\label{eq:photo_loss}
\end{equation}
where $\hat{\mathbf{C}}_i$ is the rendered color from Eq.~\ref{eq:rendering_color} and $\mathbf{c}_i$ is the ground truth.

\paragraph{Eikonal Regularization.}
Following~\cite{yariv2021volume, gropp2020implicitgeometricregularizationlearning}, the Eikonal loss enforces unit gradient norm for SDF regularization:
\begin{equation}
\mathcal{L}_{\text{eik}} = 
\frac{1}{M} \sum_{j=1}^{M} 
\big(\|\nabla f(\mathbf{x}_j)\|_2 - 1\big)^2.
\label{eq:eikonal_loss}
\end{equation}

\paragraph{Specular Regularization.}
To prevent overfitting of the specular term in Eqs.~\ref{eq:specular}--\ref{eq:total_color}:
\begin{equation}
\mathcal{L}_{\text{spec}} = 
\frac{1}{N} \sum_{i=1}^{N} 
\big\| \mathbf{c}_s^{(i)} \big\|_2^2.
\label{eq:specular_loss}
\end{equation}

\paragraph{Entropy Regularization.}
To avoid sharp discontinuities in volume opacity, we include an entropy-based regularizer:
\begin{equation}
\mathcal{L}_{\text{entropy}} = 
- w_i \log(w_i) - (1 - w_i) \log(1 - w_i),
\label{eq:entropy_loss}
\end{equation}
where $w_i$ denotes the per-point rendering weight.

\paragraph{Total Objective.}
The final objective is defined as:
\begin{equation}
\mathcal{L}_{\text{total}} =
\lambda_{\text{photo}}\mathcal{L}_{\text{photo}} +
\lambda_{\text{eik}}\mathcal{L}_{\text{eik}} +
\lambda_{\text{spec}}\mathcal{L}_{\text{spec}} +
\lambda_{\text{entropy}}\mathcal{L}_{\text{entropy}}.
\label{eq:total_loss}
\end{equation}

\subsection{Stage 2: Surface Mesh Refinement}
\label{sec:stage2}

After Stage~1 converges, we extract a coarse mesh $\mathbf{M}_{\text{coarse}}$ 
from the learned density field using the Marching Cubes algorithm~\cite{lorensen1987marchingcubes}.  
This mesh, along with the refined camera poses $\mathbf{P}_i^{\text{refined}}$ (Eq.~\ref{eq:pose_refinement}), 
initializes Stage~2 for fine surface and appearance refinement.  
We adopt a differentiable mesh refinement framework similar to NeRF2Mesh~\cite{tang2022nerf2mesh}, 
which combines differentiable rasterization~\cite{Laine2020diffrast} with photometric supervision to jointly optimize mesh geometry and appearance.  
Unlike NeRF2Mesh, our refinement starts from a pose-corrected implicit field, 
allowing more stable optimization and improved texture alignment across views.

\paragraph{Appearance Refinement.}
We employ \textit{nvdiffrast}~\cite{Laine2020diffrast} for differentiable rendering.  
The extracted mesh is rasterized, and per-pixel color gradients are computed by interpolating vertex features into the image space.  
Since the appearance network learned in Stage~1 already encodes both diffuse and specular components, 
we reuse these priors in Stage~2 to accelerate convergence and maintain consistent photometric quality.  
The pixel-wise photometric loss $\mathcal{L}_{\text{photo}}$ (Eq.~\ref{eq:photo_loss}) 
is again used to supervise the refinement of geometry and appearance jointly.

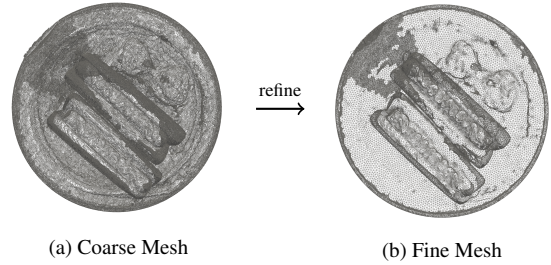
\begin{figure}[!t]
    \centering
    \begin{minipage}{0.98\linewidth} 
        \hspace*{-6pt} 
        \begin{tikzpicture}[baseline=(current bounding box.center)]
            \node (coarse) at (0,0)
                {\includegraphics[width=0.40\textwidth]{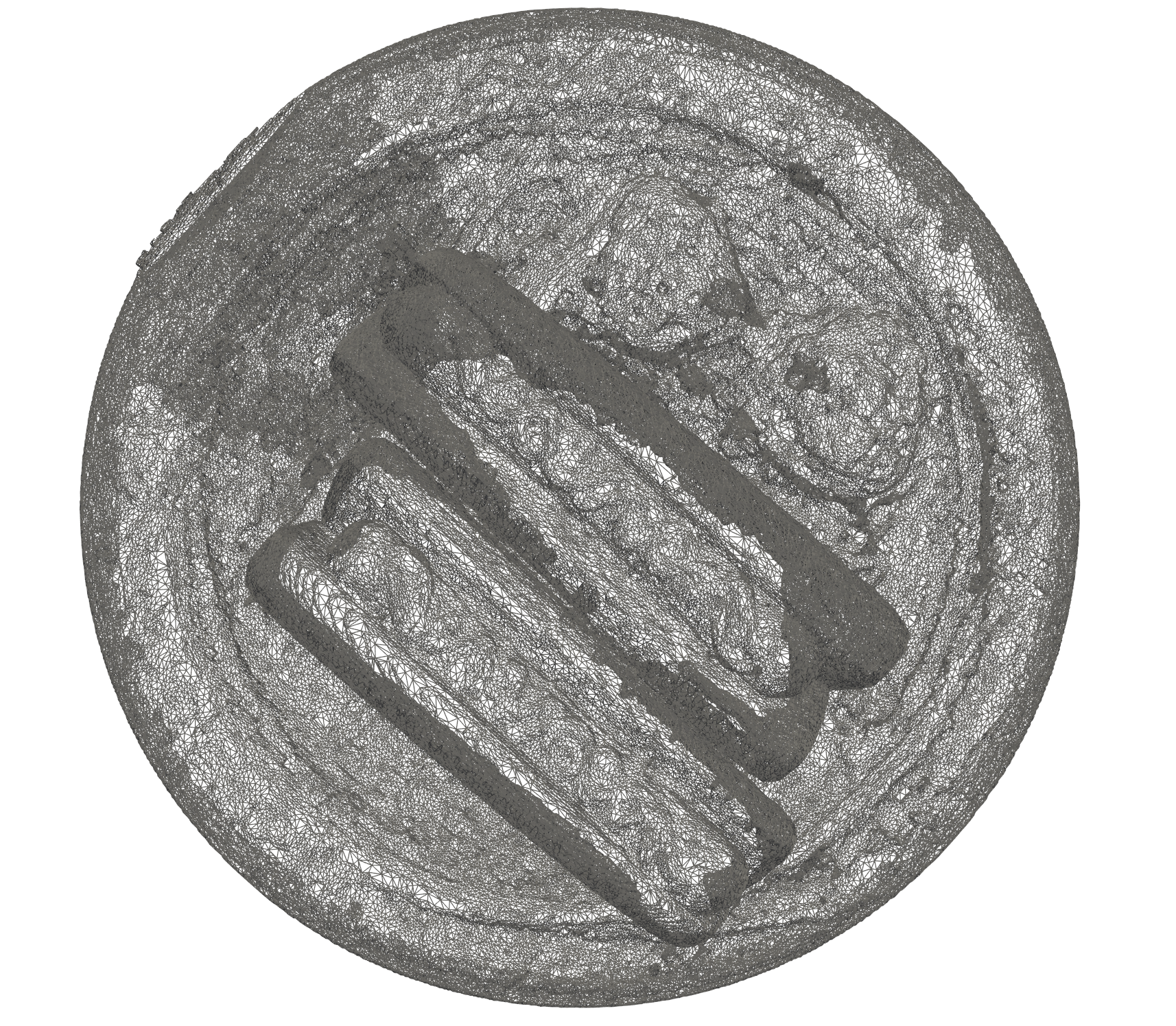}};
            \node (fine) at (4.3,0)
                {\includegraphics[width=0.40\textwidth]{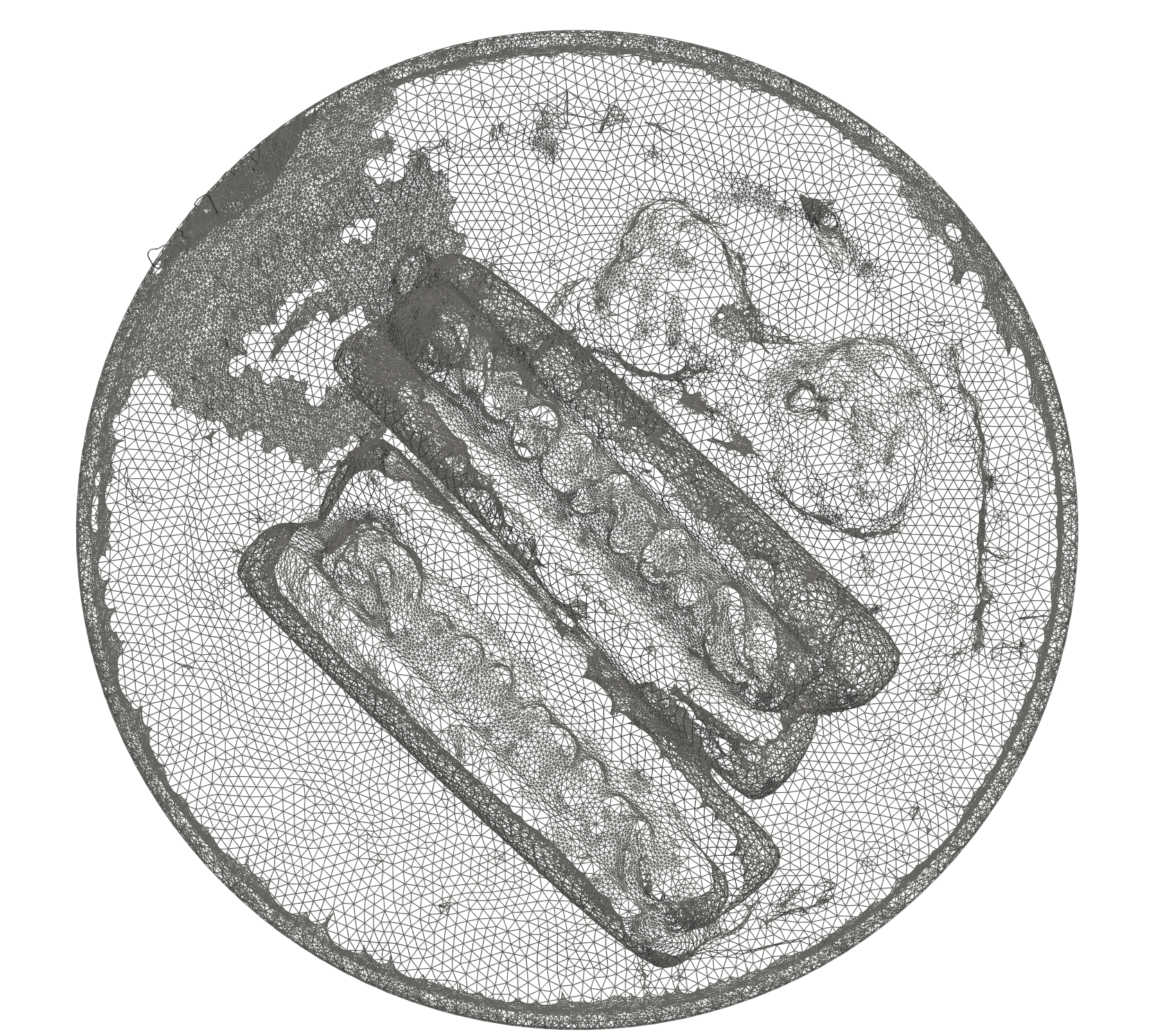}};
            \draw[->, thick, shorten >=2pt, shorten <=2pt]
                (coarse.east) -- (fine.west)
                node[midway, above, yshift=1pt]{\scriptsize refine};
            \node[below=2pt of coarse] {\footnotesize (a) Coarse Mesh};
            \node[below=2pt of fine]   {\footnotesize (b) Fine Mesh};
        \end{tikzpicture}
    \end{minipage}
    \caption{The coarse mesh contains over-dense regions and surface irregularities, while the refined mesh shows adaptive subdivision and decimation, producing a smoother and more efficient representation.}
    \label{fig:mesh_refinement}
\end{figure}

\begin{figure}[!t]
    \centering
    \hspace*{-15pt}
    \begin{tabular}{cc}
        \includegraphics[width=0.25\textwidth]{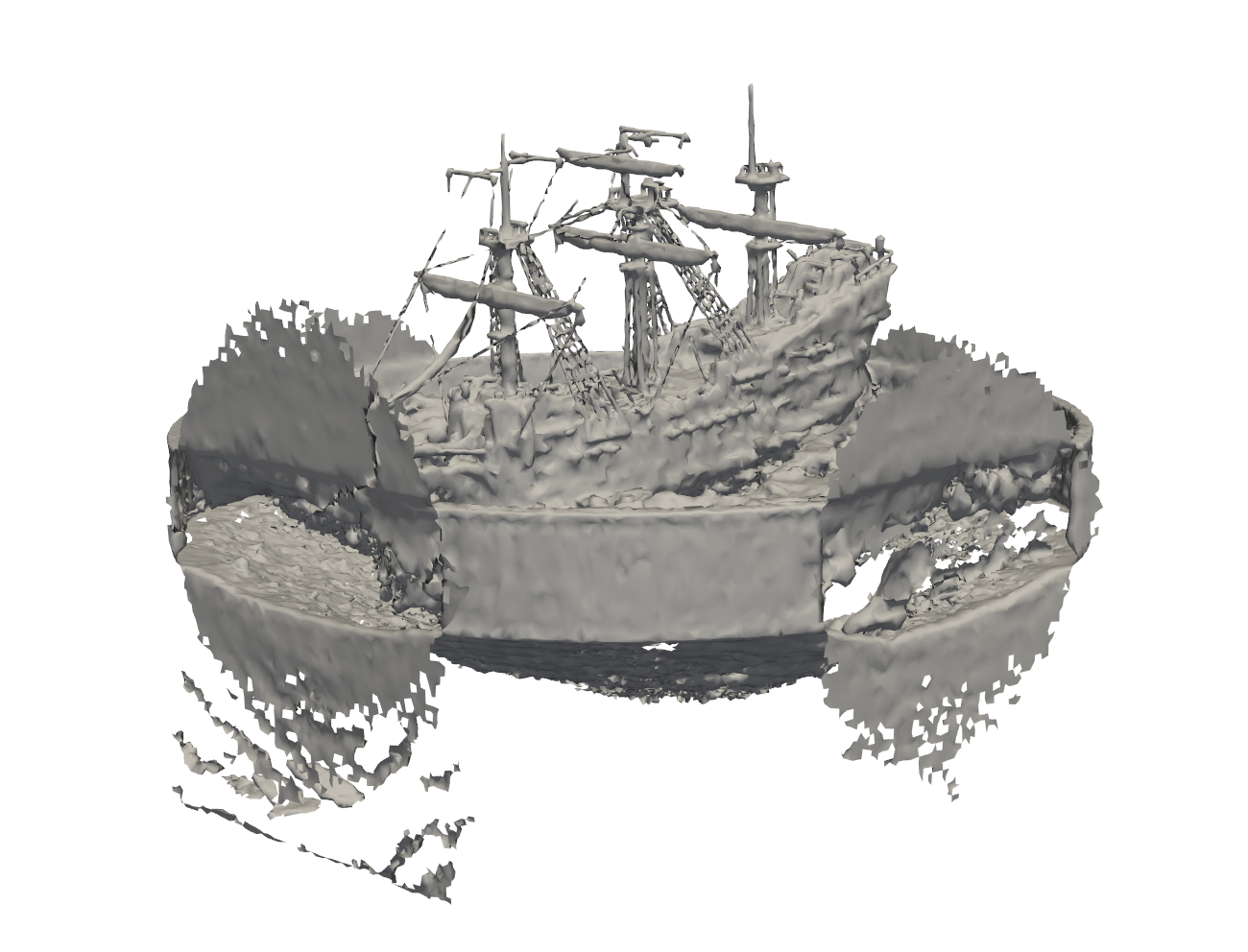} &
        \includegraphics[width=0.25\textwidth]{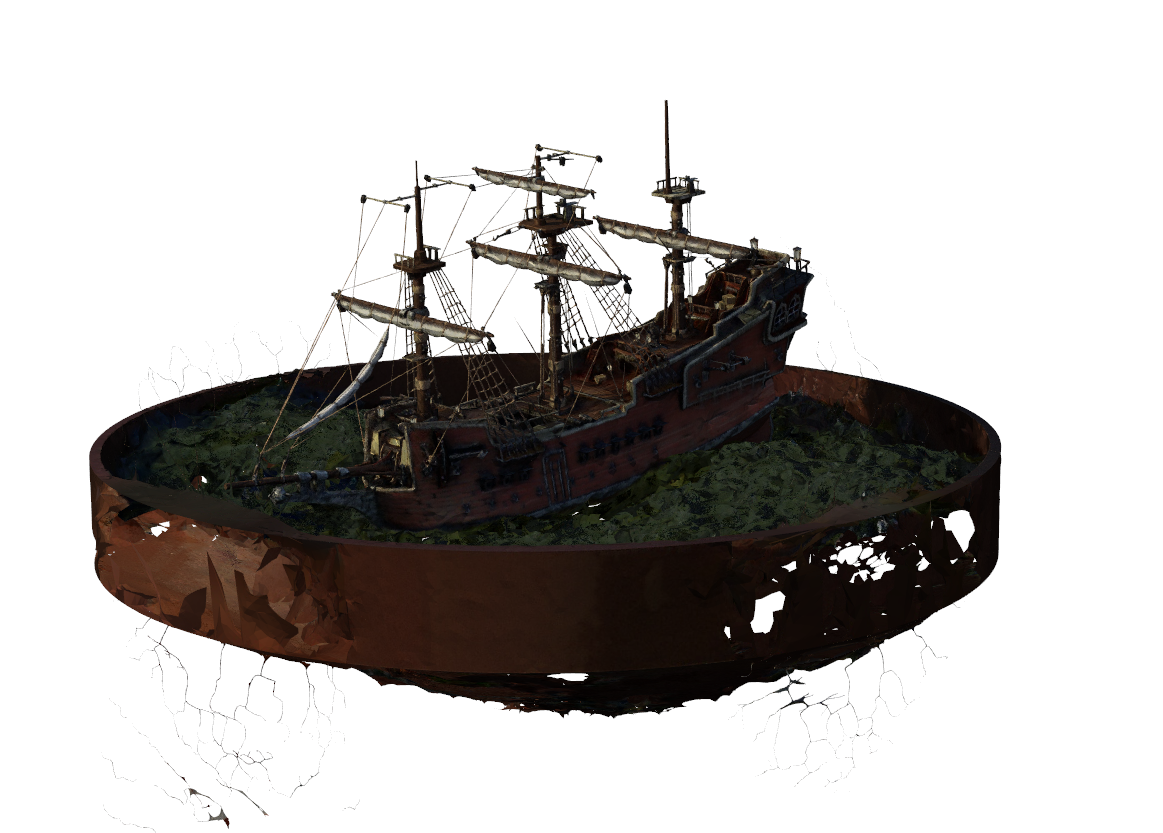} \\
        \footnotesize (a) Coarse Mesh & \footnotesize (b) Fine Mesh \\
    \end{tabular}
    \caption{Comparison between coarse and fine mesh representations. The coarse mesh captures the overall geometry, while the fine mesh provides detailed surface information.}
    \label{fig:mesh_comparison}
\end{figure}

\paragraph{Iterative Mesh Optimization.}
The coarse mesh $\mathbf{M}_{\text{coarse}} = \{V, F\}$ is often over-dense and may contain surface irregularities.  
We assign a trainable offset $\Delta v_i$ to each vertex $v_i \in V$, 
and optimize these offsets via backpropagation through the differentiable renderer.  
Following the iterative refinement strategy introduced in~\cite{tang2022nerf2mesh}, 
we adaptively adjust face density based on accumulated rendering errors.  
Faces with large reprojection error are subdivided to capture fine structures, 
while low-error regions are decimated to reduce redundancy.  
This process is repeated over multiple iterations until convergence. As shown in Figure \ref{fig:mesh_refinement} and \ref{fig:mesh_comparison}, the refinement progressively smooths irregularities and reduces redundancy, producing a more accurate and efficient mesh representation.

\paragraph{Loss Regularization.}
To ensure geometric smoothness and avoid over-deformation, 
we apply Laplacian regularization and vertex offset penalties:
\begin{equation}
\mathcal{L}_{\text{smooth}} =
\frac{1}{|S_i|}\sum_{j \in S_i} 
\| (v_i + \Delta v_i) - (v_j + \Delta v_j) \|_2^2,
\end{equation}
\begin{equation}
\mathcal{L}_{\text{offset}} =
\sum_i \| \Delta v_i \|_2^2.
\end{equation}
The total refinement loss is defined as:
\begin{equation}
\mathcal{L}_{\text{refine}} =
\lambda_{\text{photo}}\mathcal{L}_{\text{photo}} +
\lambda_{\text{smooth}}\mathcal{L}_{\text{smooth}} +
\lambda_{\text{offset}}\mathcal{L}_{\text{offset}}.
\end{equation}

\paragraph{Texture Baking and Export.}
After refinement, we unwrap UV coordinates for the final mesh $\mathbf{M}_{\text{fine}}$ 
using \textit{xatlas}~\cite{greer2019xatlas}, a fast and robust open-source UV mapping library. We then bake the diffuse color $\mathbf{c}_d$ and specular features $\mathbf{f}_s$ into separate texture maps $\mathbf{I}_d$ and $\mathbf{I}_s$, respectively. Following the real-time rendering strategy of MobileNeRF~\cite{chen2022mobilenerf}, we integrate the learned specular MLP into a lightweight fragment shader for 
view-dependent rendering. The resulting textured mesh is fully compatible with standard 3D software such as Blender, Unity, and robotic simulation frameworks, enabling interactive visualization and downstream real-time applications.

\subsection{Training Strategies}

\subsubsection{Trainable Occupancy Grid}
To accelerate training and improve sampling efficiency, we maintain a dynamic trainable occupancy grid following~\cite{mueller2022instant}.  
The grid estimates per-cell occupancy probabilities, adaptively guiding ray sampling toward geometry-dense regions while skipping empty space.  
It is optimized using a mean-squared error loss:
\begin{equation}
\mathcal{L}_{\text{grid}} =
\text{MSE}\!\left(
\mathbf{D}_{\text{pred}}[M], \,
\mathbf{D}_{\text{target}}[M]
\right),
\label{eq:grid_loss}
\end{equation}
where $M$ denotes the set of valid sampled locations.  
This adaptive grid greatly reduces redundant ray queries, resulting in faster convergence without compromising reconstruction quality.

\subsubsection{Coarse-to-Fine Positional Encoding}
We adopt a coarse-to-fine feature scheduling strategy inspired by BARF \cite{lin2021barf} to improve training stability under pose uncertainty. Rather than exposing high-frequency features at the early stages, we progressively introduce finer-scale encodings throughout training. To extend this concept to hash-grid features, we build upon the multi-resolution hash-grid positional encoding framework proposed in BAA-NGP \cite{liu2023baangp} and interpolate between feature levels as:
\begin{equation}
\gamma_k(\mathbf{x}; \alpha) =
w_k(\alpha) \, d_k(\mathbf{x}) +
(1 - w_k(\alpha)) \, d_{\alpha}(\mathbf{x}),
\label{eq:coarse_to_fine}
\end{equation}
where $d_k(\mathbf{x})$ denotes the feature at level $k$, and $w_k(\alpha)$ is a progressive cosine window defined as:
\begin{equation}
w_k(\alpha) =
\begin{cases}
0, & \alpha < k,\\[3pt]
\frac{1 - \cos((\alpha - k)\pi)}{2}, & 0 \le \alpha - k < 1,\\[3pt]
1, & \alpha - k \ge 1,
\end{cases}
\label{eq:window_func}
\end{equation}
with $\alpha \in [0,1]$ representing normalized training progress.  
This gradual activation of frequency bands prevents early overfitting to noisy poses and allows high-frequency geometric and photometric details to emerge smoothly as training progresses.

\section{Implementation Details}

We adopt a two-stage training pipeline based on a grid-based neural representation.  
For scene encoding, we utilize a multi-resolution hash grid~\cite{mueller2022instant} with $L=16$ levels, where the grid resolution increases geometrically from $N_{\text{min}} = 14$ to $N_{\text{max}} = 4069$.  
This hierarchical encoding efficiently captures both coarse and fine spatial details across multiple scales.

\paragraph{Training Setup.}  
The network is trained for $30{,}000$ iterations using two independent AdamW optimizers one for the NeRF parameters and another for the camera pose parameters.  
The NeRF optimizer is initialized with a learning rate of $1\times10^{-3}$, exponentially decaying to $1\times10^{-5}$, while the pose optimizer uses a lower learning rate that decays from $1\times10^{-4}$ to $1\times10^{-6}$.  
To improve stability, we apply a coarse-to-fine (C2F) smooth positional encoding on the hash grids, gradually revealing higher-resolution features over the normalized training interval $[0.1, 0.5]$.  
This strategy prevents early overfitting to high-frequency noise and promotes smooth convergence of both geometry and camera pose parameters. A summary of key training hyperparameters used in our experiments is provided in Table~\ref{tab:training_params}.

\paragraph{Stage 1.}  
In the first stage, we train the grid-based NeRF model jointly with camera pose refinement.  
A coarse surface mesh is extracted from the learned signed distance field using the Marching Cubes algorithm~\cite{lorensen1987marchingcubes} at a resolution of $512^3$, with a density threshold of $0.001$.  
This mesh, along with the refined camera poses, serves as initialization for the second stage.

\paragraph{Stage 2.}  
During the refinement stage, geometry and appearance are fine-tuned using differentiable rendering, initialized from the outputs of Stage~1.  
To simulate real-world scenarios, we perturb the ground-truth camera poses by adding Gaussian noise $\mathcal{N}(0, 0.15\mathbf{I})$, following the protocol in~\cite{lin2021barf}. A trainable density grid is maintained throughout training to accelerate ray marching, following the occupancy-grid update strategy from Instant-NGP~\cite{mueller2022instant}.

\paragraph{Framework and Hardware.}  
Our architecture is implemented on top of the torch-ngp~\cite{torch-ngp} framework.  
To enable joint optimization of camera pose and scene parameters, we extend the system with gradient flow support that allows backpropagation through camera parameters.  
This is achieved by modifying the $\mathrm{SE}(3)$ pose parameterization and integrating custom backward passes within the torch-ngp training pipeline.  
All experiments are conducted on a single NVIDIA RTX~4070 GPU with 12~GB of VRAM.

\begin{table}[t]
\centering
\caption{Summary of key training hyperparameters used in both stages of RePose-NeRF.}
\label{tab:training_params}
\begin{tabular}{lc}
\hline
\textbf{Parameter} & \textbf{Value} \\
\hline
Learning rate (NeRF) & $1\times10^{-3} \rightarrow 1\times10^{-5}$ \\
Learning rate (Pose) & $1\times10^{-4} \rightarrow 1\times10^{-6}$ \\
Iterations & 30k \\
Levels ($L$) & 16 \\
Resolution range & 14–4069 \\
Noise level $\mathcal{N}(0,0.15\mathbf{I})$ & Gaussian \\
\hline
\end{tabular}
\end{table}
\FloatBarrier

\section{Experimental Results and Evaluation}

We conduct comprehensive experiments to evaluate the performance of RePose-NeRF in terms of pose refinement, novel view synthesis quality, and mesh reconstruction accuracy. 
All experiments are performed on standard NeRF benchmarks to ensure fair comparison and reproducibility.

\subsection{Experimental Setup}

We evaluate RePose-NeRF on two widely used datasets: LLFF and Blender (NeRF-Synthetic).  
The \{LLFF dataset contains forward-facing real-world scenes captured under natural illumination and non-ideal pose configurations, used to evaluate photorealism and robustness under real-world conditions.  
The Blender (NeRF-Synthetic) dataset consists of eight synthetic scenes with ground-truth poses and geometry, providing a controlled environment for quantitative assessment of both reconstruction and rendering quality.

\begin{figure}[!t]
    \centering
    \setlength{\tabcolsep}{1pt}
    \renewcommand{\arraystretch}{0.8}
    \begin{tabular}{cccc}
        \includegraphics[width=0.115\textwidth]{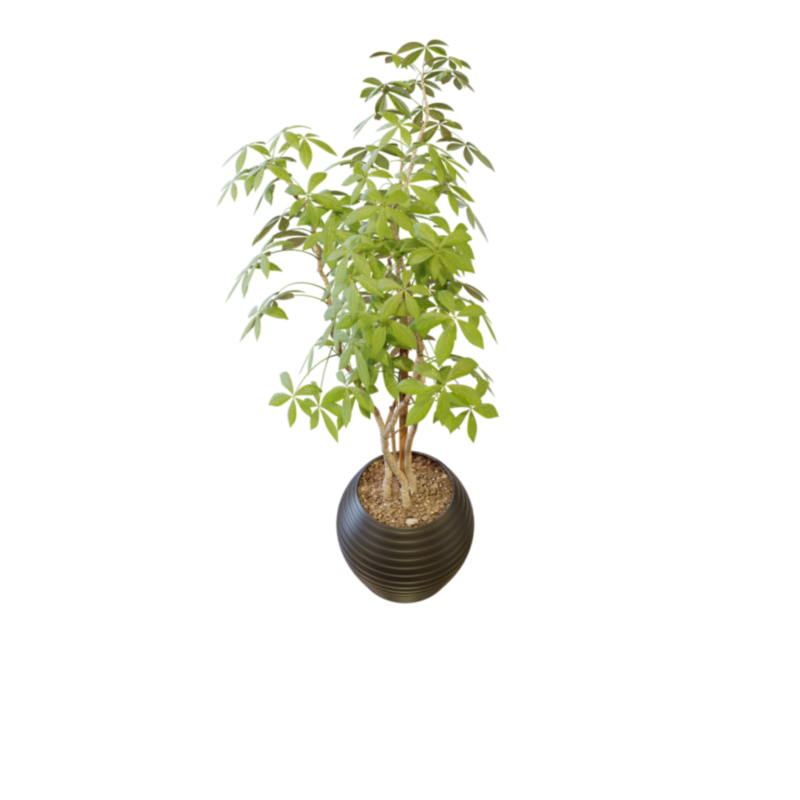} &
        \includegraphics[width=0.115\textwidth]{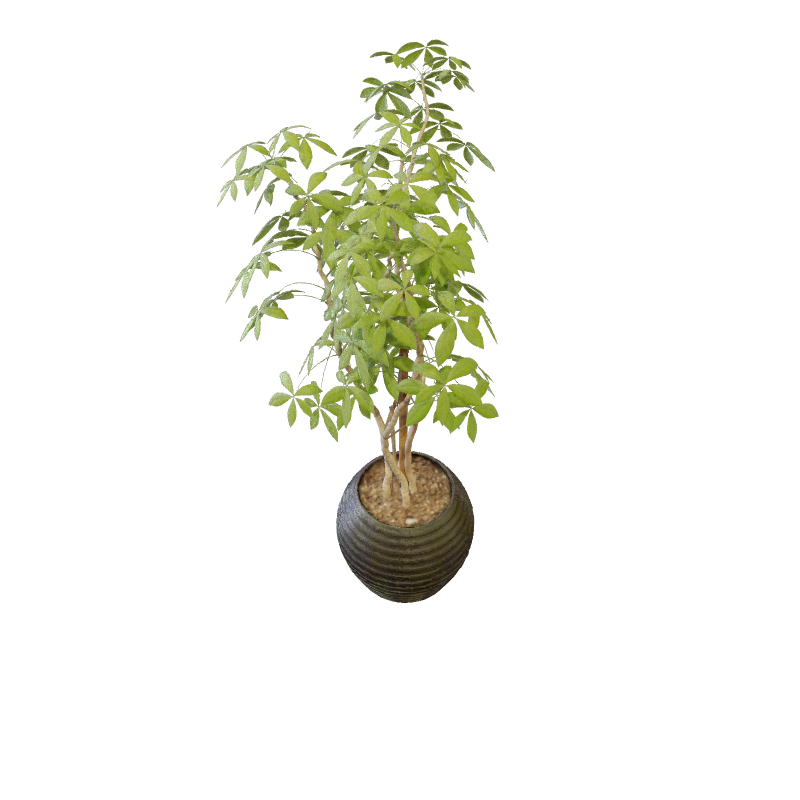} &
        \includegraphics[width=0.115\textwidth]{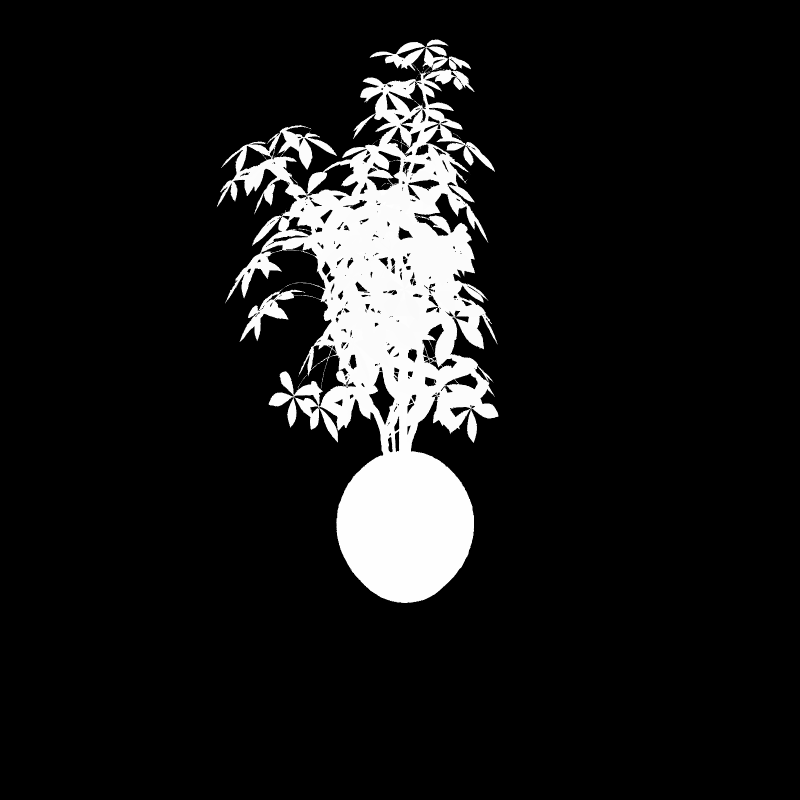} &
        \includegraphics[width=0.115\textwidth]{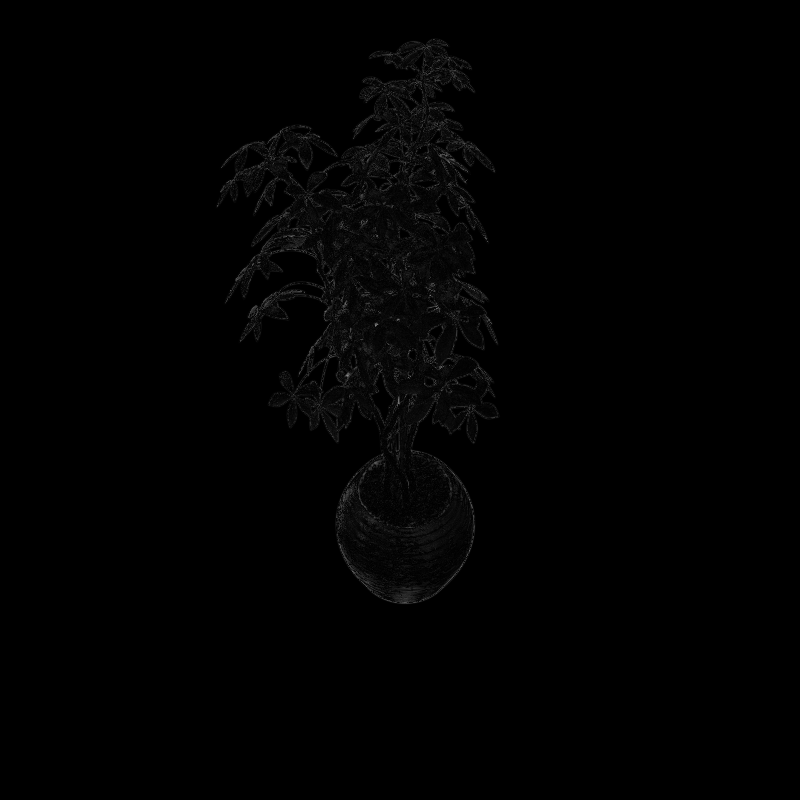} \\[-1pt]
        \includegraphics[width=0.115\textwidth]{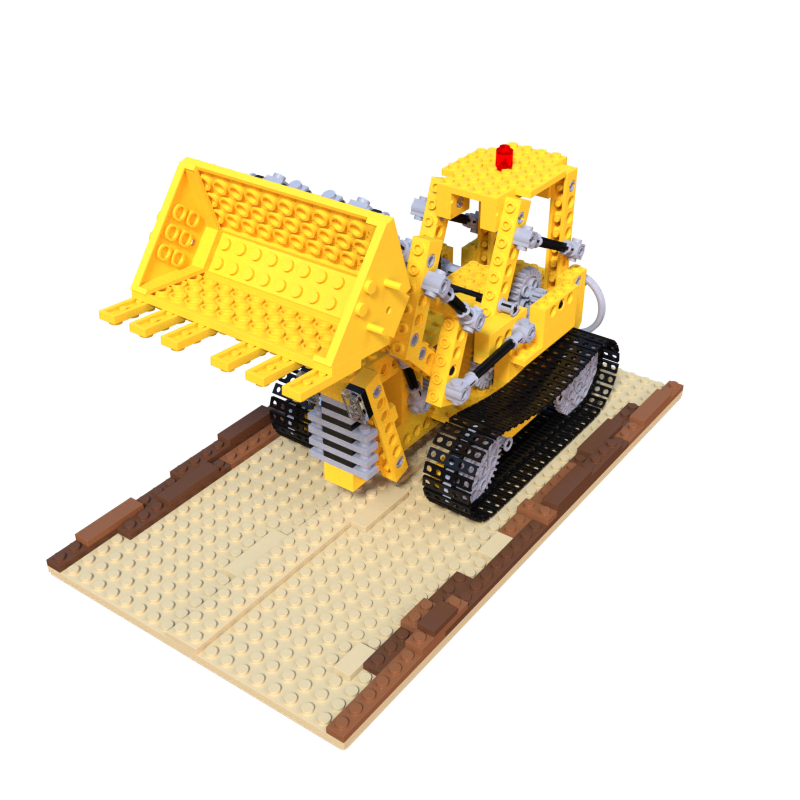} &
        \includegraphics[width=0.115\textwidth]{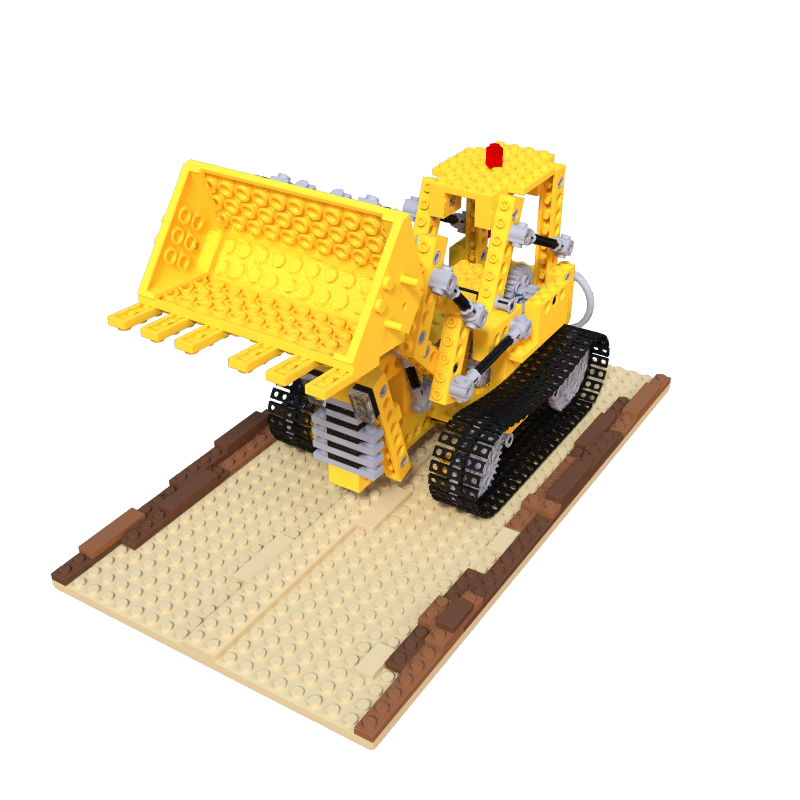} &
        \includegraphics[width=0.115\textwidth]{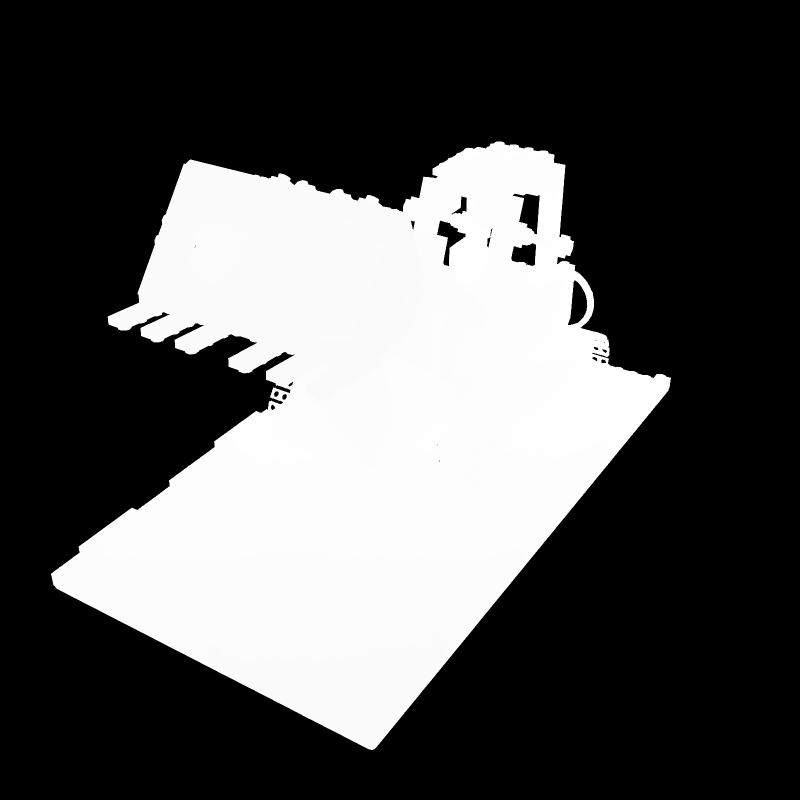} &
        \includegraphics[width=0.115\textwidth]{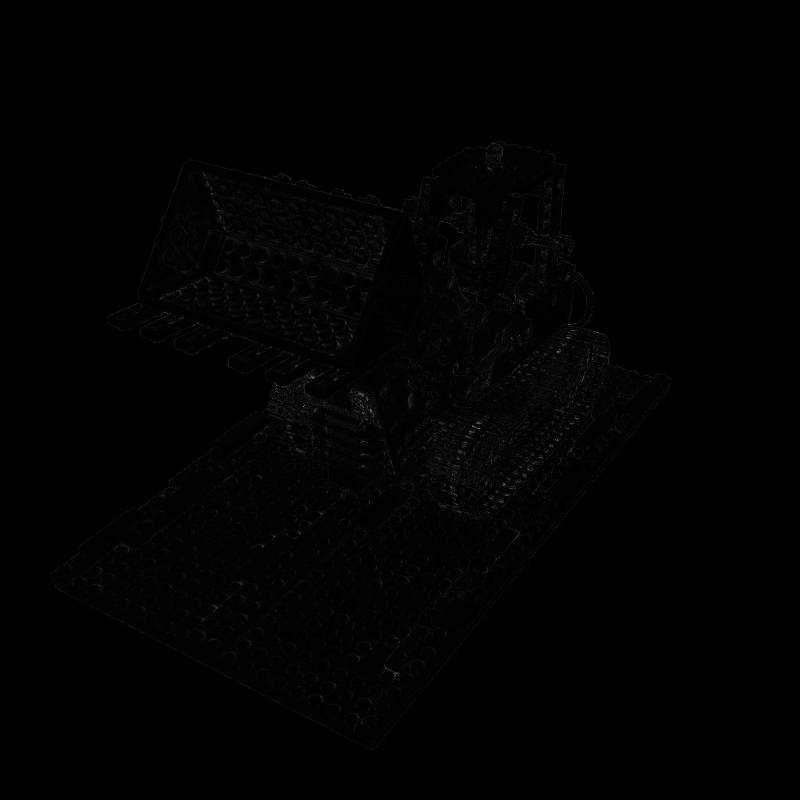} \\[-1pt]
        \includegraphics[width=0.115\textwidth]{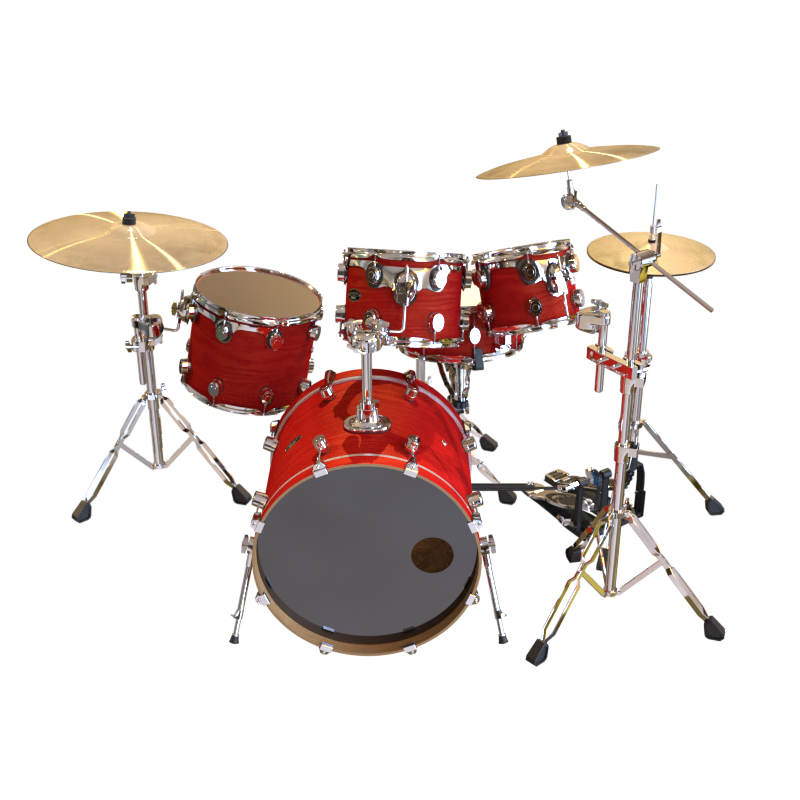} &
        \includegraphics[width=0.115\textwidth]{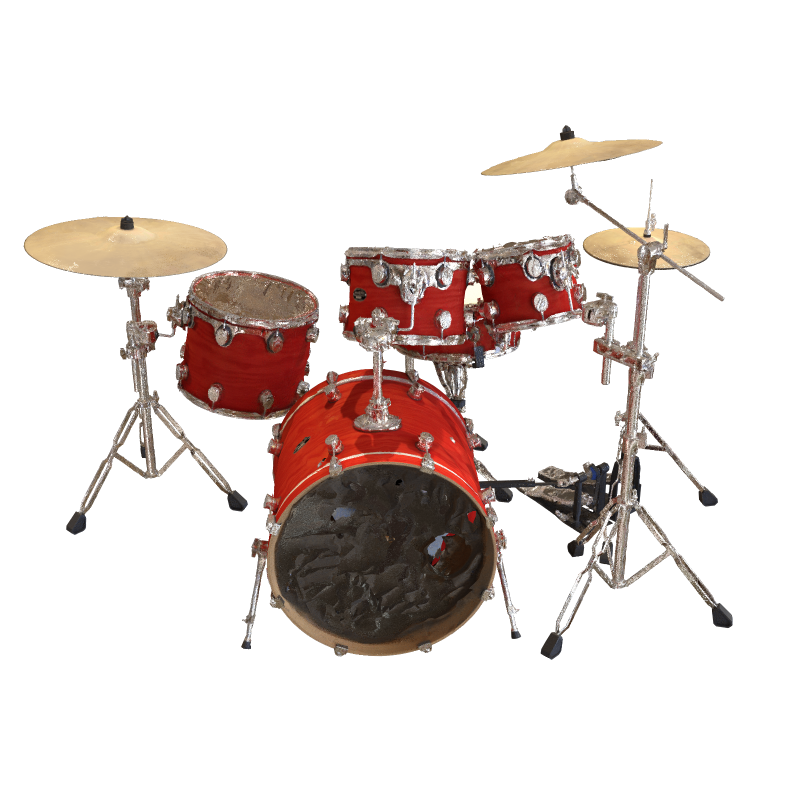} &
        \includegraphics[width=0.115\textwidth]{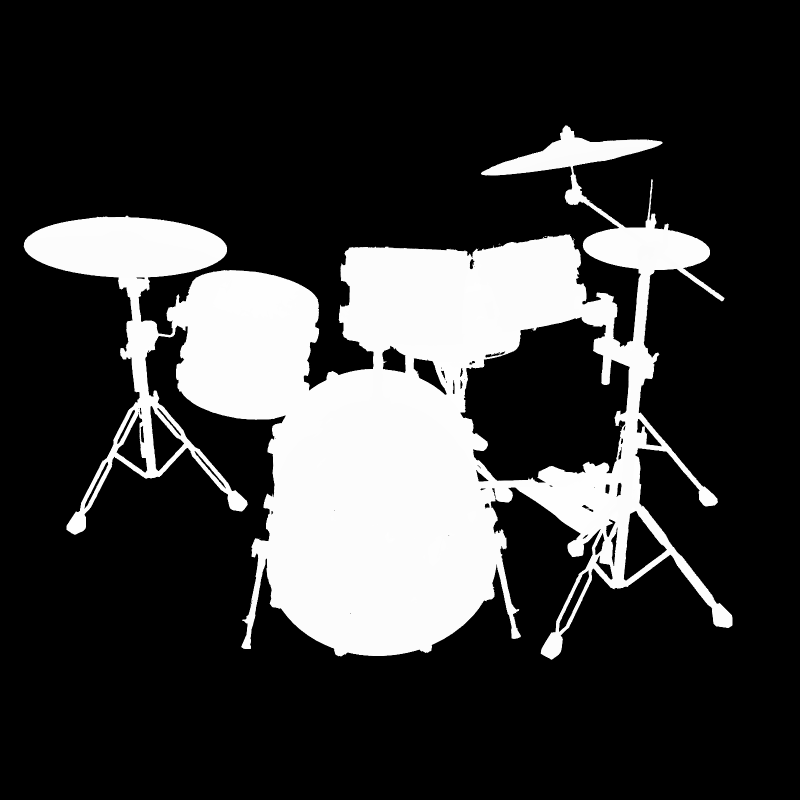} &
        \includegraphics[width=0.115\textwidth]{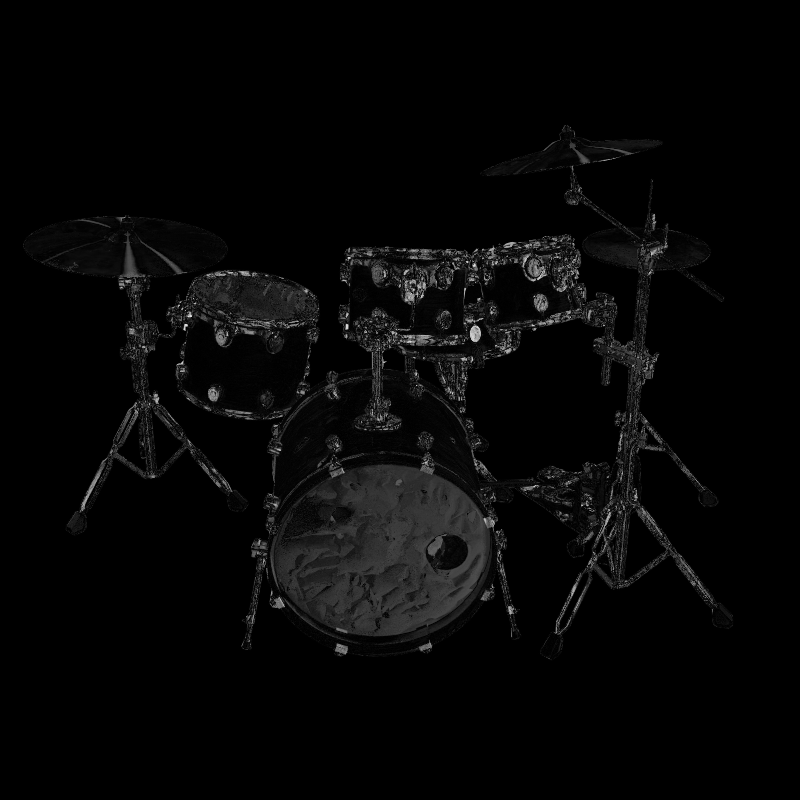} \\[-1pt]
        \includegraphics[width=0.115\textwidth]{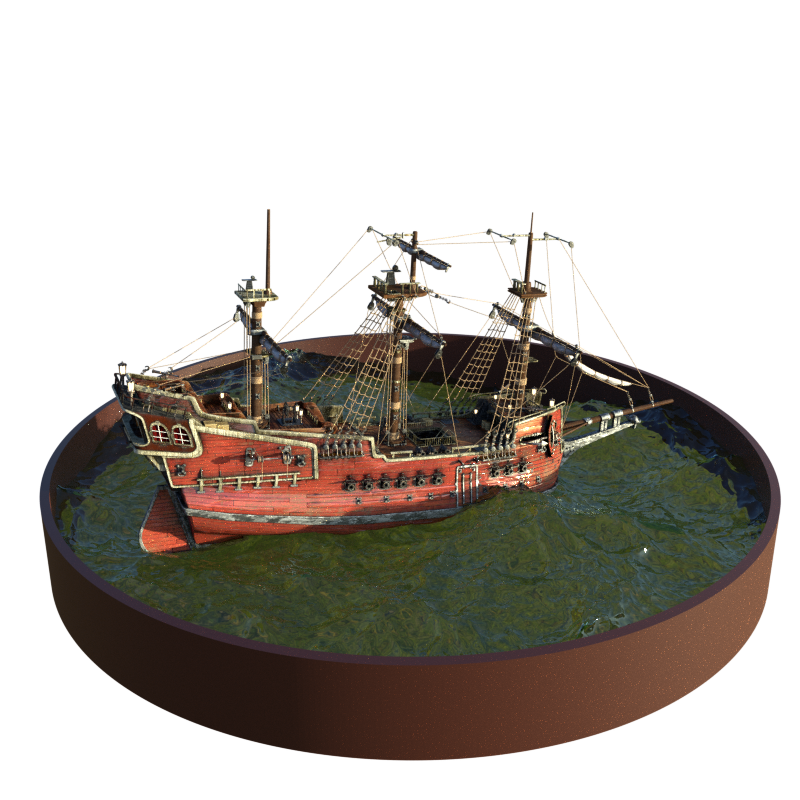} &
        \includegraphics[width=0.115\textwidth]{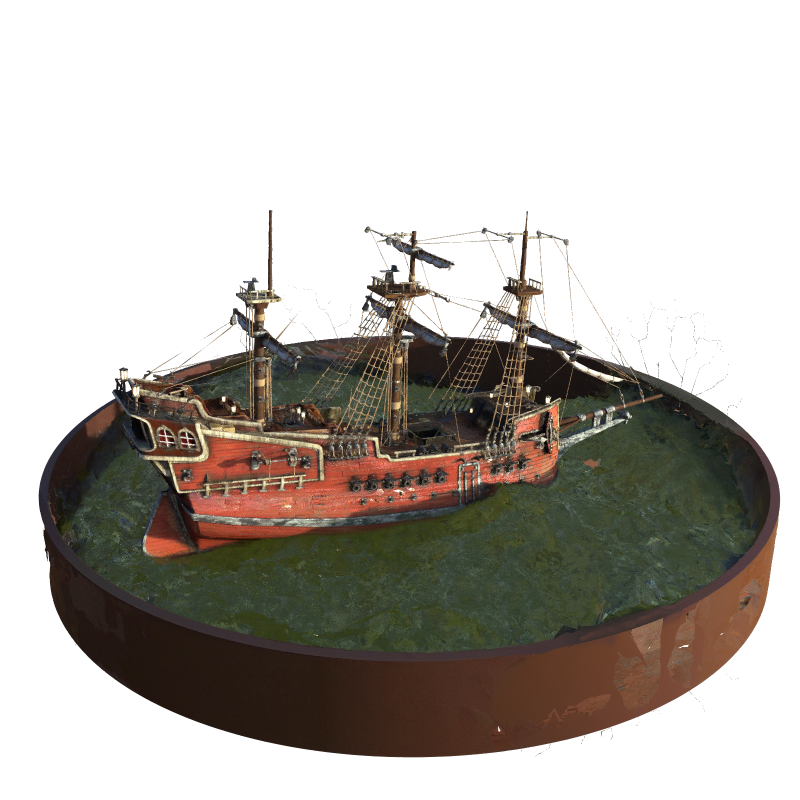} &
        \includegraphics[width=0.115\textwidth]{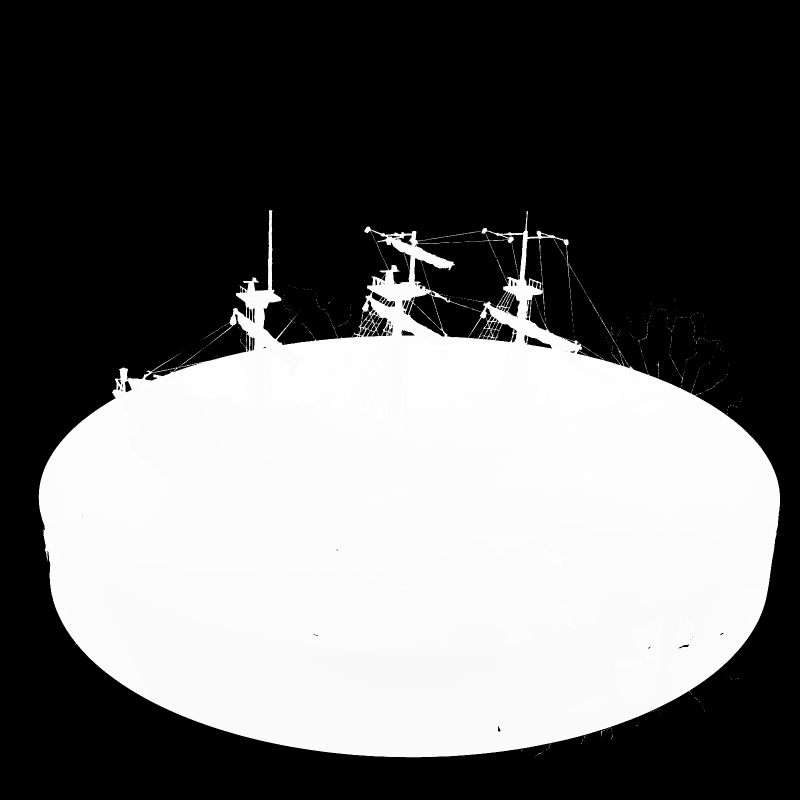} &
        \includegraphics[width=0.115\textwidth]{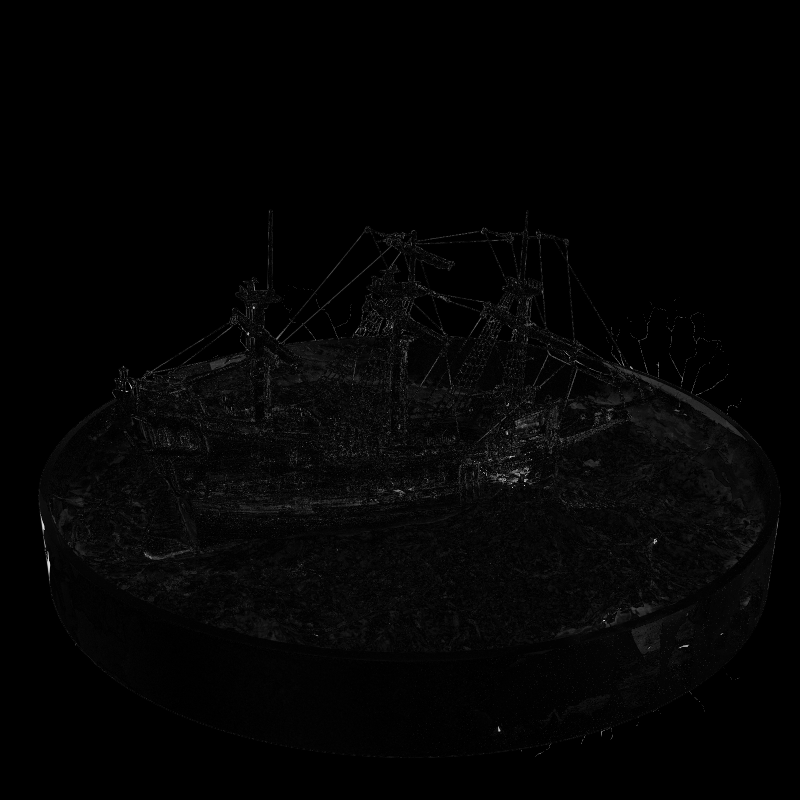} \\[3pt]
        \footnotesize GT & \footnotesize Pred & \footnotesize Depth & \footnotesize Error Depth \\
    \end{tabular}
    \vspace{-4pt}
    \caption{Qualitative visualization of rendering results for four NeRF-Synthetic classes (\textit{Ficus, Lego, Drums, Ship}). Each column shows the GT, predicted RGB render, estimated depth, and depth error map.}
    \label{fig:nerf_render_four_class}
    \vspace{-4pt}
\end{figure}

To simulate real-world noise, we follow the same perturbation protocol as BARF~\cite{lin2021barf}, where the ground-truth camera poses are modified by adding by Gaussian noise $\mathcal{N}(0, 0.15I)$.  
We benchmark our method against BARF~\cite{lin2021barf}, which jointly optimizes camera poses and scene parameters during training.  
The evaluation covers three key aspects:

\begin{itemize}[leftmargin=*]
    \item \textbf{Pose Refinement:} Accuracy of the refined extrinsic parameters is measured by comparing them with ground-truth poses from the Blender dataset and LLFF (forward-facing) dataset, quantifying the ability of RePose-NeRF to recover accurate camera trajectories under noisy initialization.
    
    \item \textbf{Novel View Synthesis Quality:} Rendering quality is evaluated using PSNR, MS-SSIM, and LPIPS metrics to assess photorealism, perceptual similarity, and structural consistency. (see Figure \ref{fig:nerf_render_four_class} and \ref{fig:llff_render_four_scenes} for qualitative results)
    
    \item \textbf{Mesh Reconstruction Quality:} Geometric accuracy is evaluated using the Bidirectional Chamfer Distance (CD) between the reconstructed mesh and the ground-truth mesh, measuring surface fidelity and fine-detail preservation. For NeRF-Synthetic, where ground-truth meshes are unavailable, we cast rays from all test camera viewpoints and accumulate approximately 2.5 million samples to form a point cloud for evaluation.
\end{itemize}

\subsection{Pose Refinement and View Synthesis Quality}

Quantitative results for pose refinement and novel view synthesis are reported in Tables~\ref{tab:blender_pose} and~\ref{tab:llff_pose}. 
For the Blender dataset, the initial camera poses are perturbed by Gaussian noise $\mathcal{N}(0, 0.15I)$, following BARF~\cite{lin2021barf}. 
Compared to BARF, our proposed RePose-NeRF achieves more accurate pose refinement while delivering superior rendering fidelity across all scenes. 
On LLFF, RePose-NeRF also outperforms BARF in both photometric and perceptual metrics with significantly reduced training time. 

\begin{table*}[!t]
\centering
\scriptsize
\setlength{\tabcolsep}{3.5pt}
\renewcommand{\arraystretch}{0.9}
\caption{Quantitative evaluation on the Blender (NeRF-Synthetic) dataset. RePose-NeRF achieves improved pose accuracy and view-synthesis quality compared to BARF, with significantly reduced training time.}
\label{tab:blender_pose}
\resizebox{\textwidth}{!}{
\begin{tabular}{l||cc|cc||cc|cc|cc|cc||cc}
\toprule
\multirow{2}{*}{\textbf{Scene}} &
\multicolumn{4}{c||}{\textbf{Camera Pose Registration}} &
\multicolumn{8}{c||}{\textbf{Visual Synthesis Quality}} &
\multicolumn{2}{c}{\textbf{Training Time}}\\
\cmidrule(lr){2-15}
 & \multicolumn{2}{c|}{Rotation(°)↓} &
   \multicolumn{2}{c||}{Translation↓ (m)} &
   \multicolumn{2}{c|}{PSNR↑} &
   \multicolumn{2}{c|}{SSIM↑} &
   \multicolumn{2}{c|}{MS-SSIM↑} &
   \multicolumn{2}{c||}{LPIPS↓} &
   \textbf{BARF (hh:mm:ss)} & \textbf{Ours (mm:ss)}\\
\cmidrule(lr){2-15}
 & BARF & Ours & BARF & Ours & BARF & Ours & BARF & Ours & BARF & Ours & BARF & Ours & &\\
\midrule
Chair      & 0.096 & \textbf{0.052} & 0.428 & \textbf{0.27} & 31.16 & \textbf{35.51} & 0.954 & \textbf{0.985} & 0.990 & \textbf{0.997} & 0.044 & \textbf{0.024} & 08:35:12 & \textbf{31:46}\\
Drums      & 0.043 & \textbf{0.028} & 0.225 & \textbf{0.11} & 23.91 & \textbf{25.24} & 0.900 & \textbf{0.926} & 0.954 & \textbf{0.966} & 0.099 & \textbf{0.088} & 08:31:29 & \textbf{25:39}\\
Ficus      & 0.085 & \textbf{0.030} & 0.474 & \textbf{0.11} & 26.26 & \textbf{31.18} & 0.937 & \textbf{0.977} & 0.974 & \textbf{0.993} & 0.058 & \textbf{0.036} & 08:56:51 & \textbf{25:01}\\
Hotdog     & 0.249 & \textbf{0.078} & 1.308 & \textbf{0.43} & 34.53 & \textbf{37.31} & 0.970 & \textbf{0.982} & 0.992 & \textbf{0.994} & 0.032 & \textbf{0.032} & 08:46:51 & \textbf{25:00}\\
Lego       & 0.082 & \textbf{0.041} & 0.291 & \textbf{0.13} & 28.33 & \textbf{33.31} & 0.927 & \textbf{0.977} & \textbf{0.981} & 0.933 & 0.050 & \textbf{0.026} & 08:47:58 & \textbf{24:12}\\
Materials  & \textbf{0.844} & 1.555 & 2.692 & \textbf{5.68} & \textbf{27.94} & 25.94 & \textbf{0.939} & 0.906 & \textbf{0.984} & 0.966 & \textbf{0.058} & 0.114 & 08:32:21 & \textbf{25:19}\\
Mic        & 0.071 & \textbf{0.045} & 0.301 & \textbf{0.21} & 31.18 & \textbf{34.43} & 0.969 & \textbf{0.984} & 0.992 & \textbf{0.995} & 0.048 & \textbf{0.026} & 08:10:32 & \textbf{26:50}\\
Ship       & \textbf{0.075} & 0.882 & \textbf{0.326} & 0.367 & 27.51 & \textbf{29.71} & 0.849 & \textbf{0.882} & 0.938 & \textbf{0.944} & 0.132 & \textbf{0.123} & 08:45:29 & \textbf{21:30}\\
\midrule
\textbf{Mean} & 0.193 & 0.339 & 0.756 & 0.913 & 28.85 & 31.58 & 0.931 & 0.952 & 0.976 & 0.974 & 0.065 & 0.058 & 08:35:57 & 25:37\\
\bottomrule
\end{tabular}
}
\end{table*}

\begin{table*}[!t]
\centering
\scriptsize
\setlength{\tabcolsep}{3.5pt}
\renewcommand{\arraystretch}{0.9}
\caption{Quantitative evaluation on the LLFF dataset. RePose-NeRF achieves improved view-synthesis quality and faster convergence compared to BARF.}
\label{tab:llff_pose}
\resizebox{\textwidth}{!}{
\begin{tabular}{l||cc|cc||cc|cc|cc|cc||cc}
\toprule
\multirow{2}{*}{\textbf{Scene}} &
\multicolumn{4}{c||}{\textbf{Camera Pose Registration}} &
\multicolumn{8}{c||}{\textbf{Visual Synthesis Quality}} &
\multicolumn{2}{c}{\textbf{Training Time}}\\
\cmidrule(lr){2-15}
 & \multicolumn{2}{c|}{Rotation(°)↓} &
   \multicolumn{2}{c||}{Translation↓ (m)} &
   \multicolumn{2}{c|}{PSNR↑} &
   \multicolumn{2}{c|}{SSIM↑} &
   \multicolumn{2}{c|}{MS-SSIM↑} &
   \multicolumn{2}{c||}{LPIPS↓} &
   \textbf{BARF (hh:mm:ss)} & \textbf{Ours (mm:ss)}\\
\cmidrule(lr){2-15}
 & BARF & Ours & BARF & Ours & BARF & Ours & BARF & Ours & BARF & Ours & BARF & Ours & &\\
\midrule
Fern      & 0.191 & \textbf{0.127} & 0.192 & \textbf{0.123} & 23.96 & \textbf{24.59} & 0.709 & \textbf{0.775} & 0.916 & \textbf{0.918} & 0.390 & \textbf{0.293} & 08:51:32 & \textbf{13:17}\\
Flower    & 0.251 & \textbf{0.161} & 0.224 & \textbf{0.091} & 24.07 & \textbf{26.99} & 0.712 & \textbf{0.821} & 0.891 & \textbf{0.937} & 0.379 & \textbf{0.215} & 09:20:09 & \textbf{12:13}\\
Fortress  & 0.479 & \textbf{0.341} & 0.364 & \textbf{0.185} & 28.86 & \textbf{29.51} & 0.816 & \textbf{0.867} & 0.950 & \textbf{0.951} & 0.266 & \textbf{0.184} & 09:10:34 & \textbf{12:36}\\
Horns     & 0.304 & \textbf{0.149} & 0.222 & \textbf{0.108} & 23.12 & \textbf{26.93} & 0.734 & \textbf{0.834} & 0.915 & \textbf{0.949} & 0.423 & \textbf{0.255} & 08:55:24 & \textbf{13:05}\\
Leaves    & 1.272 & \textbf{0.853} & 0.249 & \textbf{0.183} & 18.67 & \textbf{19.64} & 0.529 & \textbf{0.655} & 0.831 & \textbf{0.877} & 0.474 & \textbf{0.329} & 09:02:35 & \textbf{13:01}\\
Orchids   & 0.627 & \textbf{0.421} & 0.404 & \textbf{0.241} & 19.37 & \textbf{19.99} & 0.570 & \textbf{0.630} & 0.835 & \textbf{0.840} & 0.423 & \textbf{0.321} & 08:45:12 & \textbf{12:59}\\
Rooms     & \textbf{0.320} & 0.353 & \textbf{0.270} & 0.346 & \textbf{31.60} & 30.01 & \textbf{0.937} & 0.931 & \textbf{0.981} & 0.952 & 0.230 & \textbf{0.219} & 08:54:43 & \textbf{10:49}\\
T-Rex     & 1.138 & \textbf{0.755} & 0.720 & \textbf{0.453} & 22.32 & \textbf{24.65} & 0.771 & \textbf{0.892} & 0.927 & \textbf{0.958} & 0.355 & \textbf{0.234} & 09:02:34 & \textbf{10:59}\\
\midrule
\textbf{Mean} & 0.573 & 0.395 & 0.331 & 0.216 & 24.12 & 25.29 & 0.722 & 0.803 & 0.902 & 0.923 & 0.364 & 0.256 & 08:59:43 & 12:10\\

\bottomrule
\end{tabular}
}
\end{table*}

\begin{figure}[!t]
    \centering
    \setlength{\tabcolsep}{1pt}
    \renewcommand{\arraystretch}{0.8}
    \begin{tabular}{cccc}
        \includegraphics[width=0.115\textwidth]{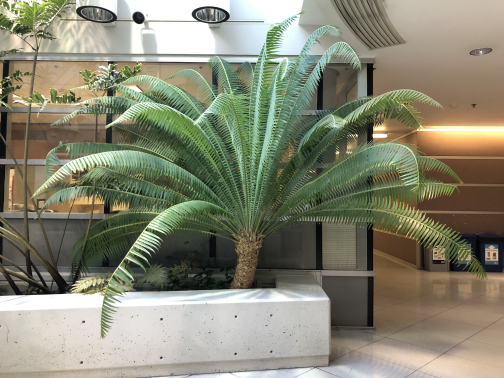} &
        \includegraphics[width=0.115\textwidth]{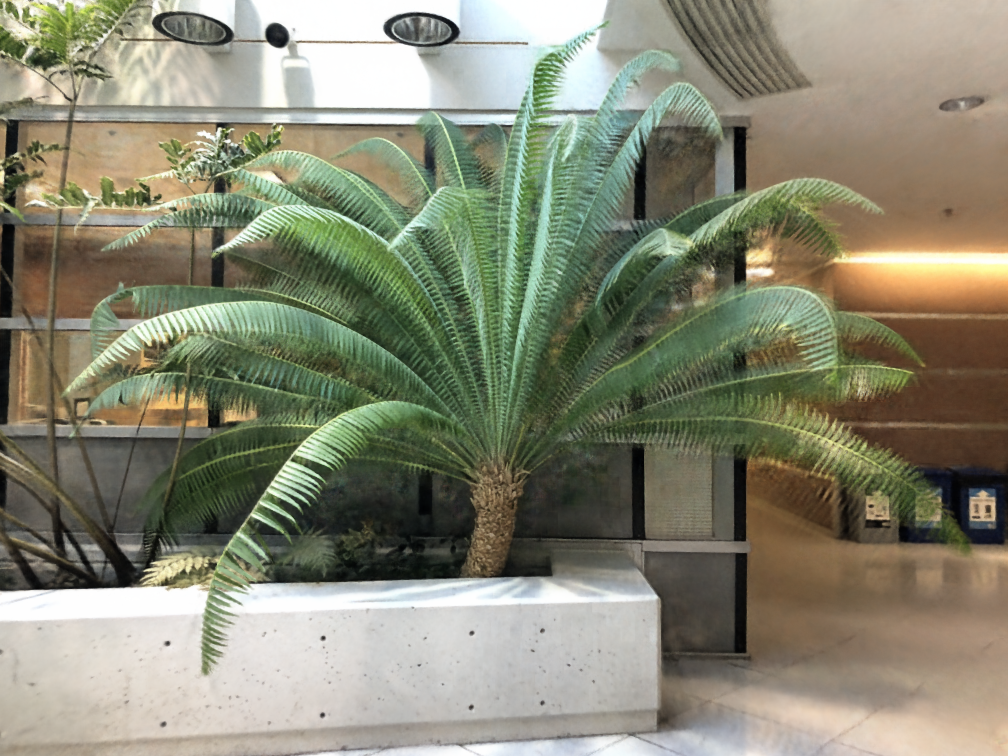} &
        \includegraphics[width=0.115\textwidth]{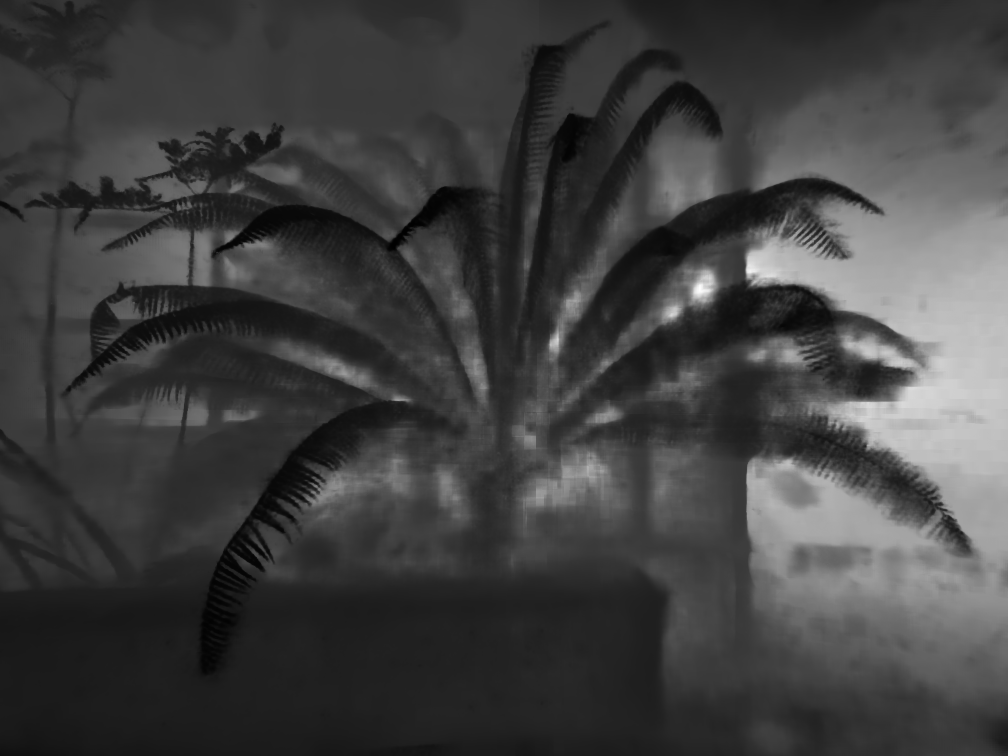} &
        \includegraphics[width=0.115\textwidth]{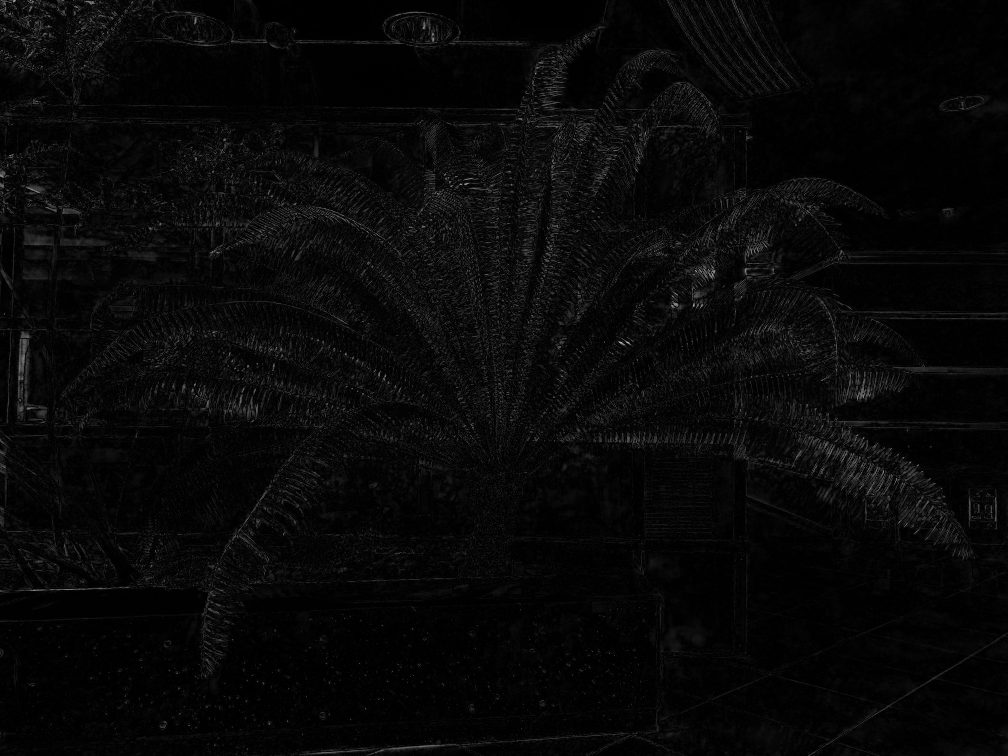} \\[-1pt]
        \includegraphics[width=0.115\textwidth]{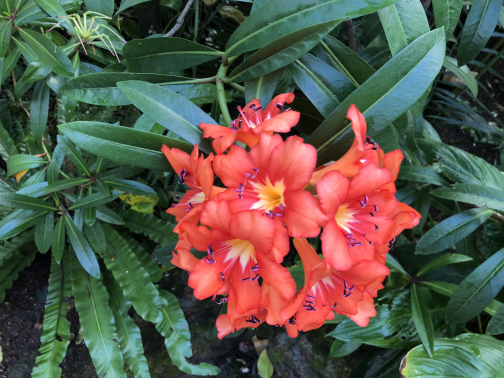} &
        \includegraphics[width=0.115\textwidth]{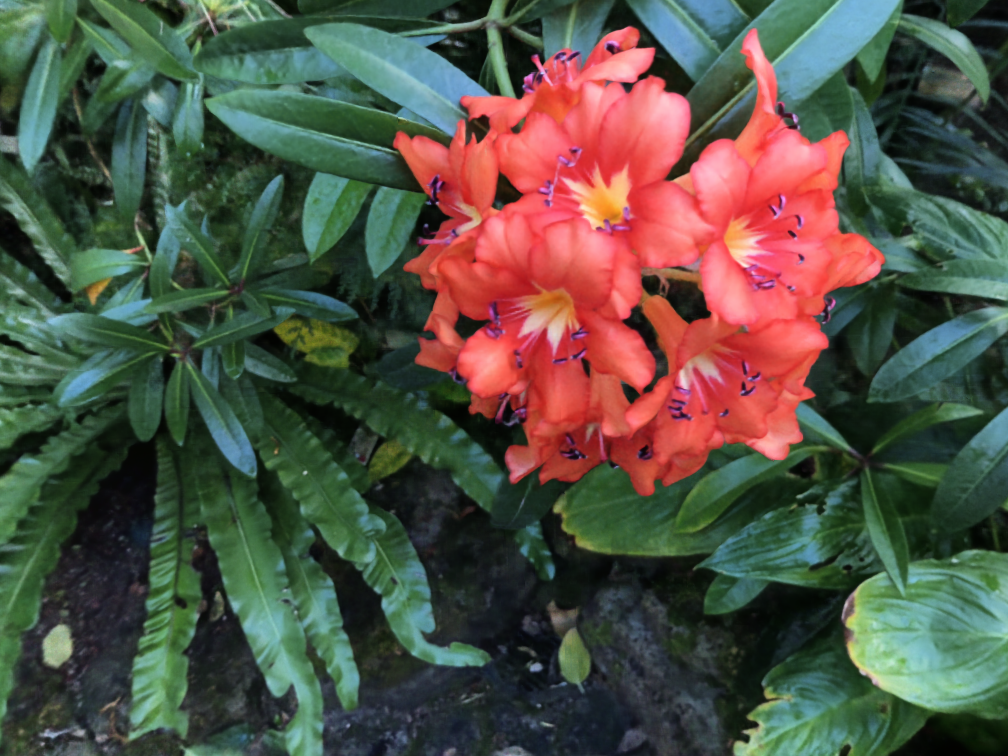} &
        \includegraphics[width=0.115\textwidth]{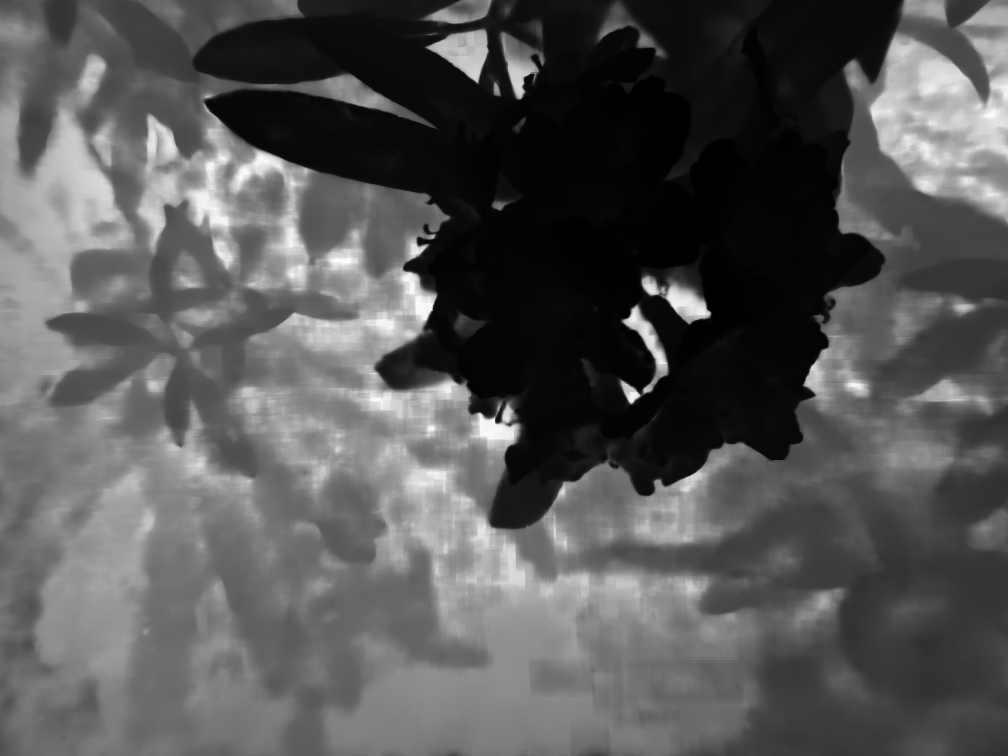} &
        \includegraphics[width=0.115\textwidth]{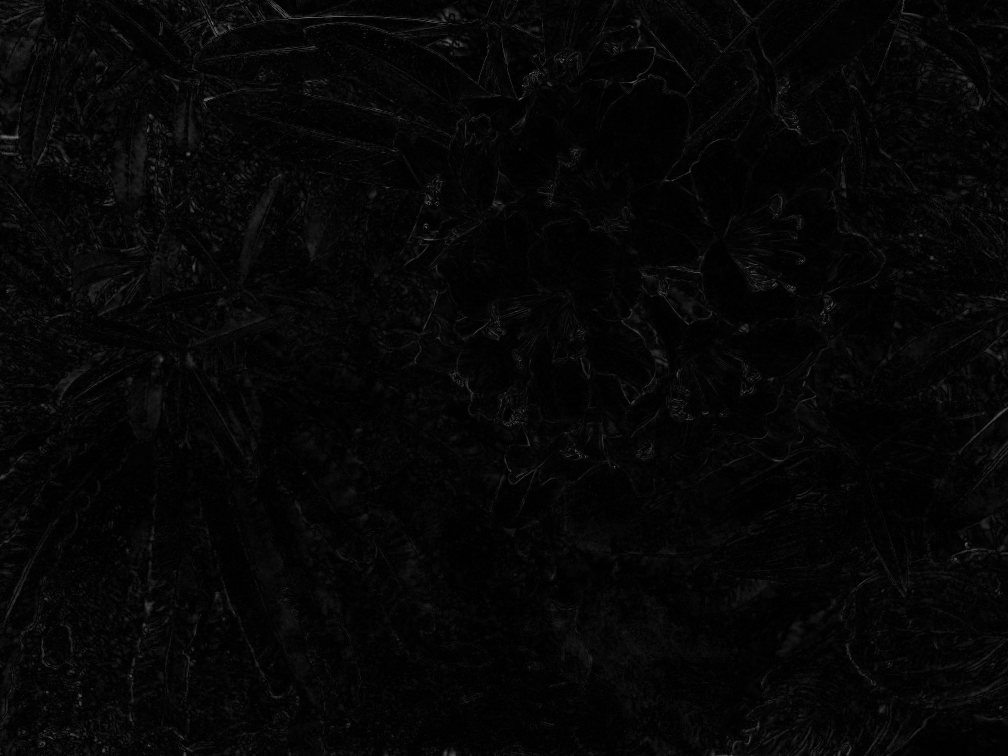} \\[-1pt]
        \includegraphics[width=0.115\textwidth]{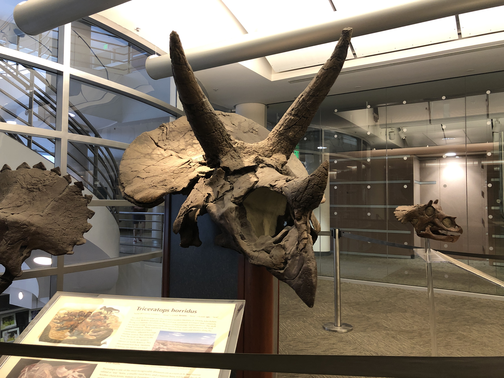} &
        \includegraphics[width=0.115\textwidth]{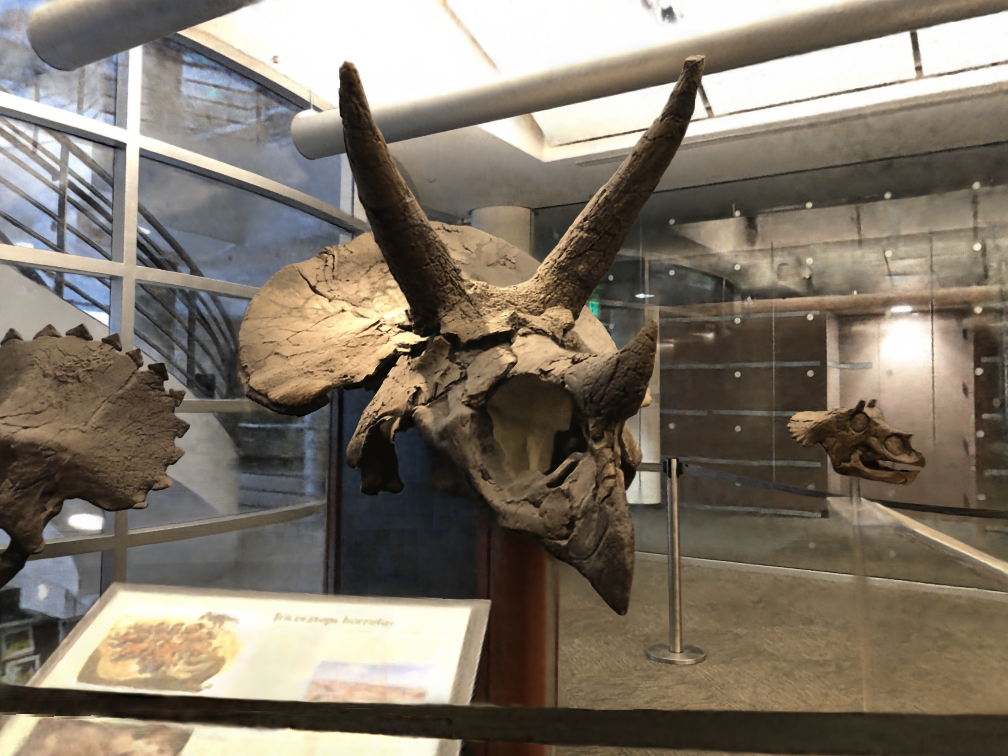} &
        \includegraphics[width=0.115\textwidth]{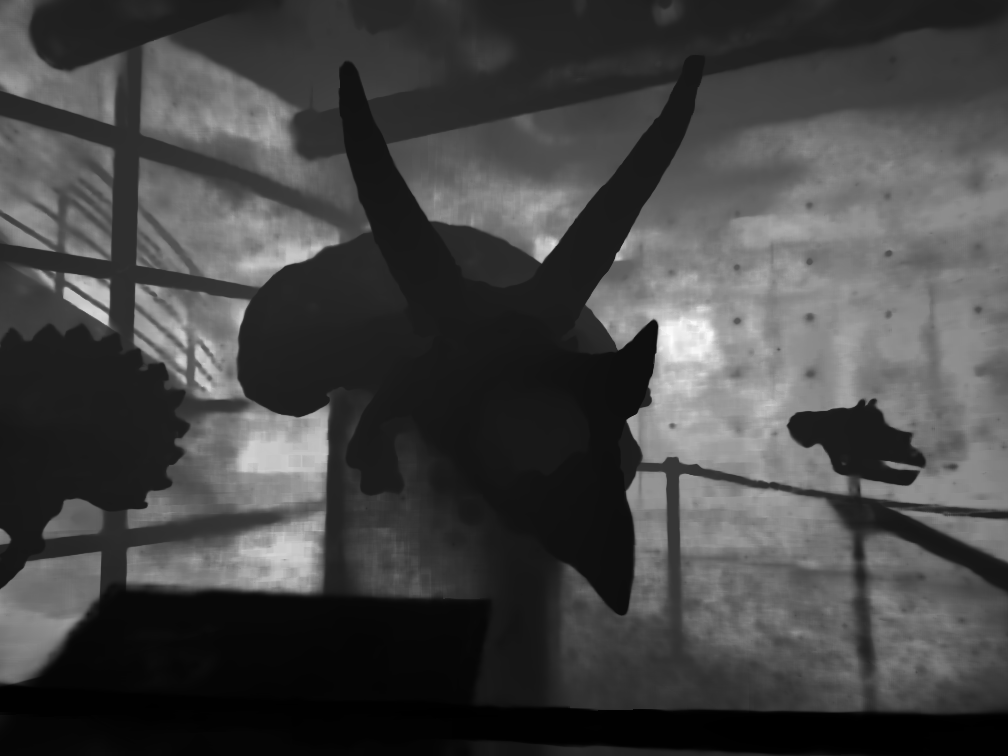} &
        \includegraphics[width=0.115\textwidth]{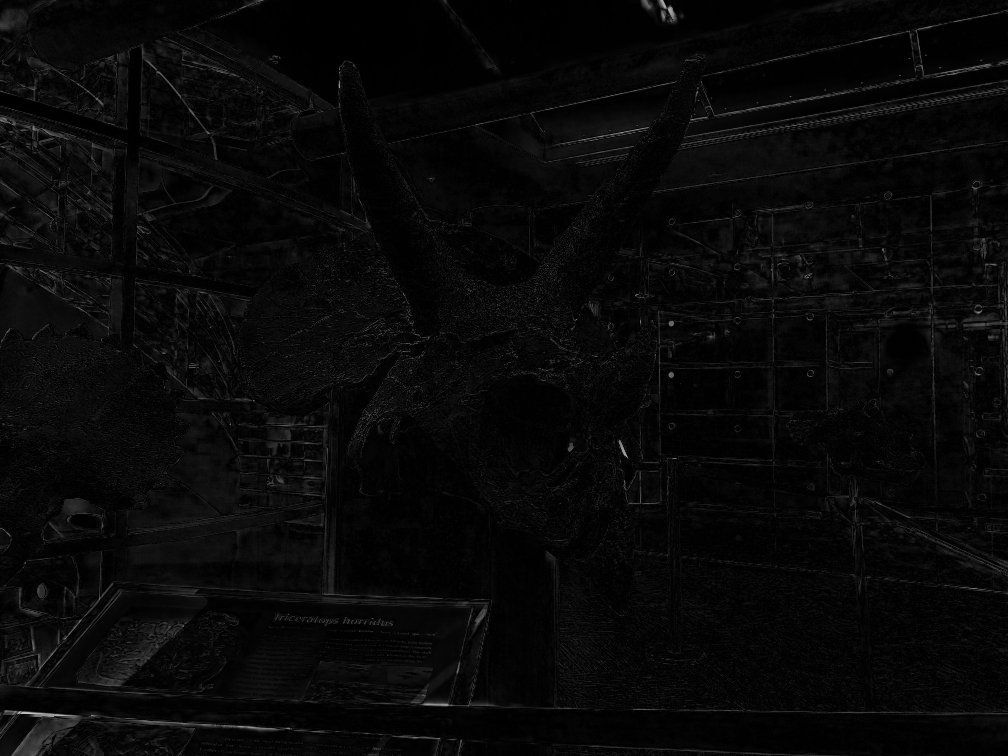} \\[-1pt]
        \includegraphics[width=0.115\textwidth]{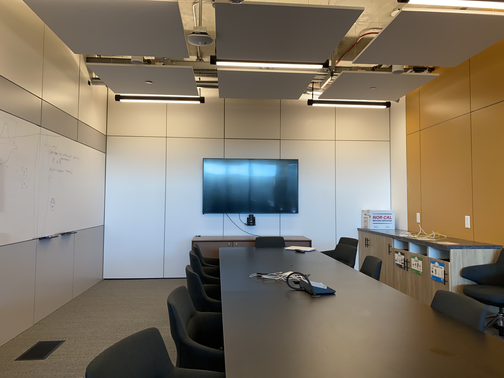} &
        \includegraphics[width=0.115\textwidth]{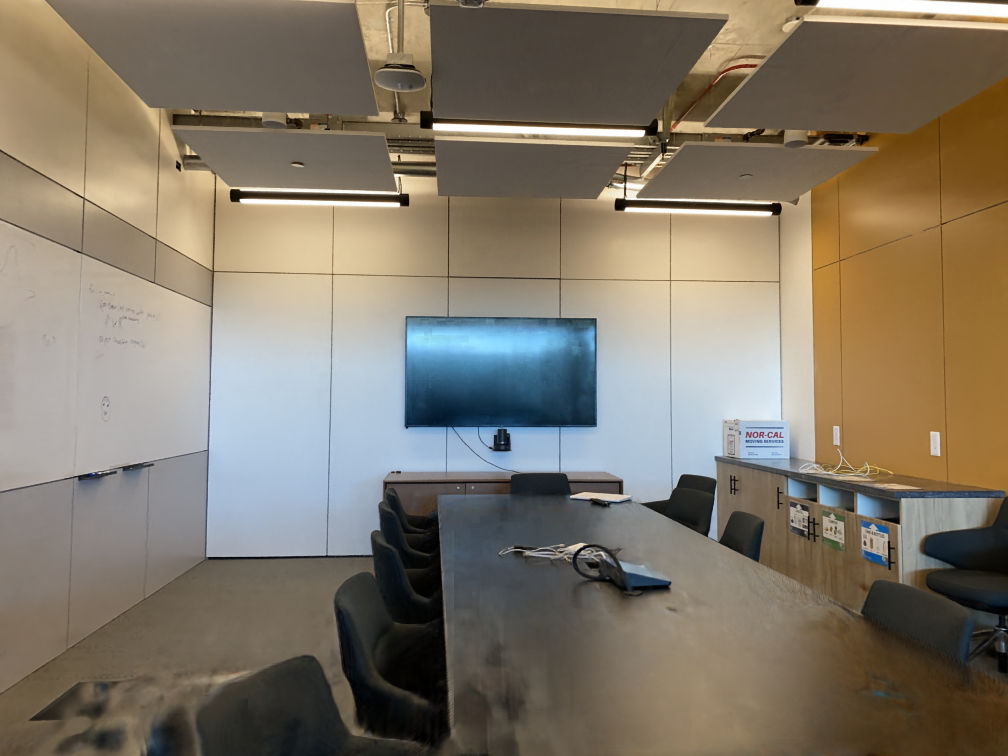} &
        \includegraphics[width=0.115\textwidth]{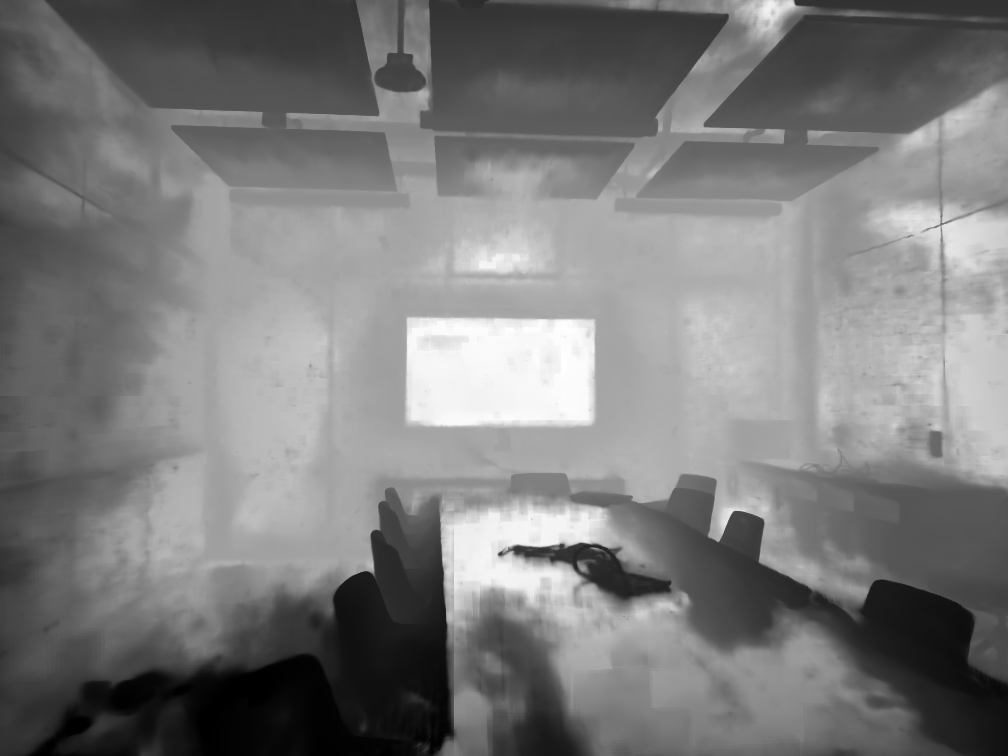} &
        \includegraphics[width=0.115\textwidth]{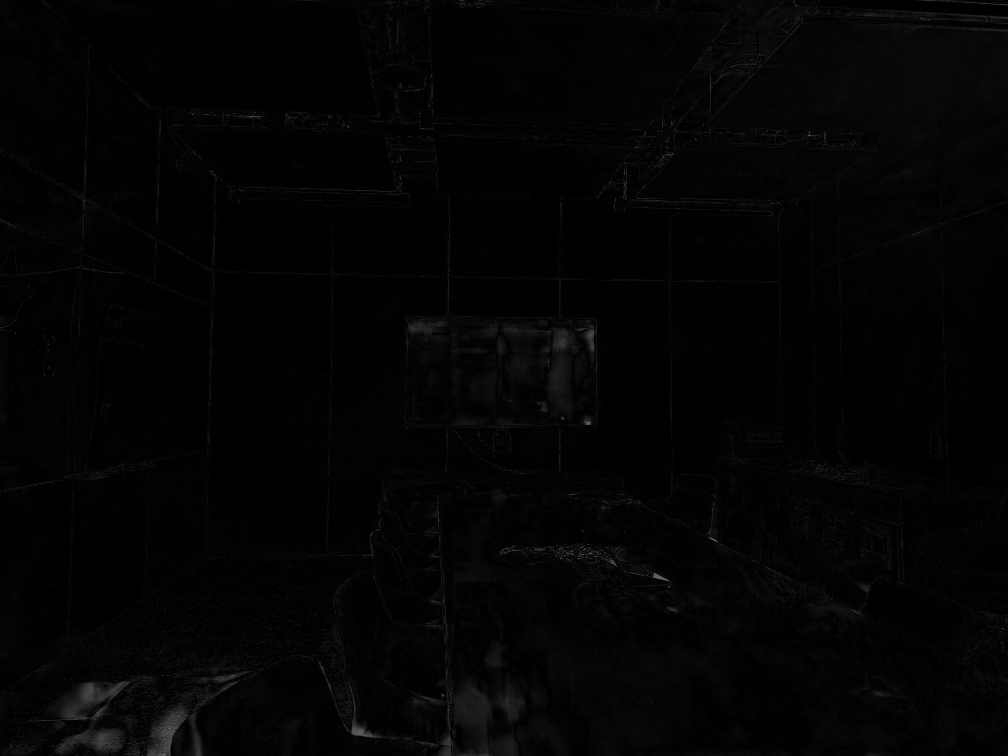} \\[3pt]
        \footnotesize GT & \footnotesize Pred & \footnotesize Depth & \footnotesize Error Depth \\
    \end{tabular}
    \vspace{-4pt}
    \caption{Qualitative visualization of rendering results for four LLFF scenes (\textit{Fern, Flower, Horns, Room}). Each column shows the ground-truth image, predicted RGB render, estimated depth, and depth error map.}
    \label{fig:llff_render_four_scenes}
    \vspace{-4pt}
\end{figure}

\subsection{Mesh Reconstruction Quality}

We further evaluate the geometric fidelity of reconstructed meshes on the NeRF-Synthetic dataset.  
Meshes are extracted using the Marching Cubes algorithm~\cite{lorensen1987marchingcubes} from the learned implicit field and compared against ground-truth surfaces using the Bidirectional Chamfer Distance (CD).  
Results (Table~\ref{tab:mesh_quality}) show that RePose-NeRF produces smoother, more accurate meshes that better capture fine structural details. Figure \ref{fig:surface_mesh_quality}  visualizes the reconstructed meshes in wireframe view, highlighting the structural accuracy and smoothness of the recovered surfaces. The wireframe renderings clearly demonstrate that RePose-NeRF captures fine geometric details with well-preserved topology.

\begin{table}[!t]
\centering
\footnotesize 
\setlength{\tabcolsep}{12pt}
\renewcommand{\arraystretch}{1.15} 
\caption{Chamfer Distance (↓) between reconstructed meshes and ground truth (GT) on Blender (NeRF-Synthetic) using \textbf{RePose-NeRF (Ours)}. Lower is better.}
\label{tab:mesh_quality}
\begin{tabular}{lc}
\toprule
\textbf{Category} & \textbf{RePose-NeRF (Ours)} \\
\midrule
Materials  & 0.0171 \\
Mic        & 0.0018 \\
Drums      & 0.0158 \\
Ficus      & 0.0039 \\
Lego       & 0.0229 \\
Hotdog     & 0.0136 \\
Chair      & 0.0068 \\
Ship       & 0.0629 \\
\midrule
\textbf{Mean} & \textbf{0.0181} \\
\bottomrule
\end{tabular}
\end{table}

\begin{figure*}[!t]
  \centering
  \setlength{\tabcolsep}{4pt}
  \renewcommand{\arraystretch}{0.8}
  \begin{tabular}{cccc}
    \includegraphics[width=0.23\textwidth]{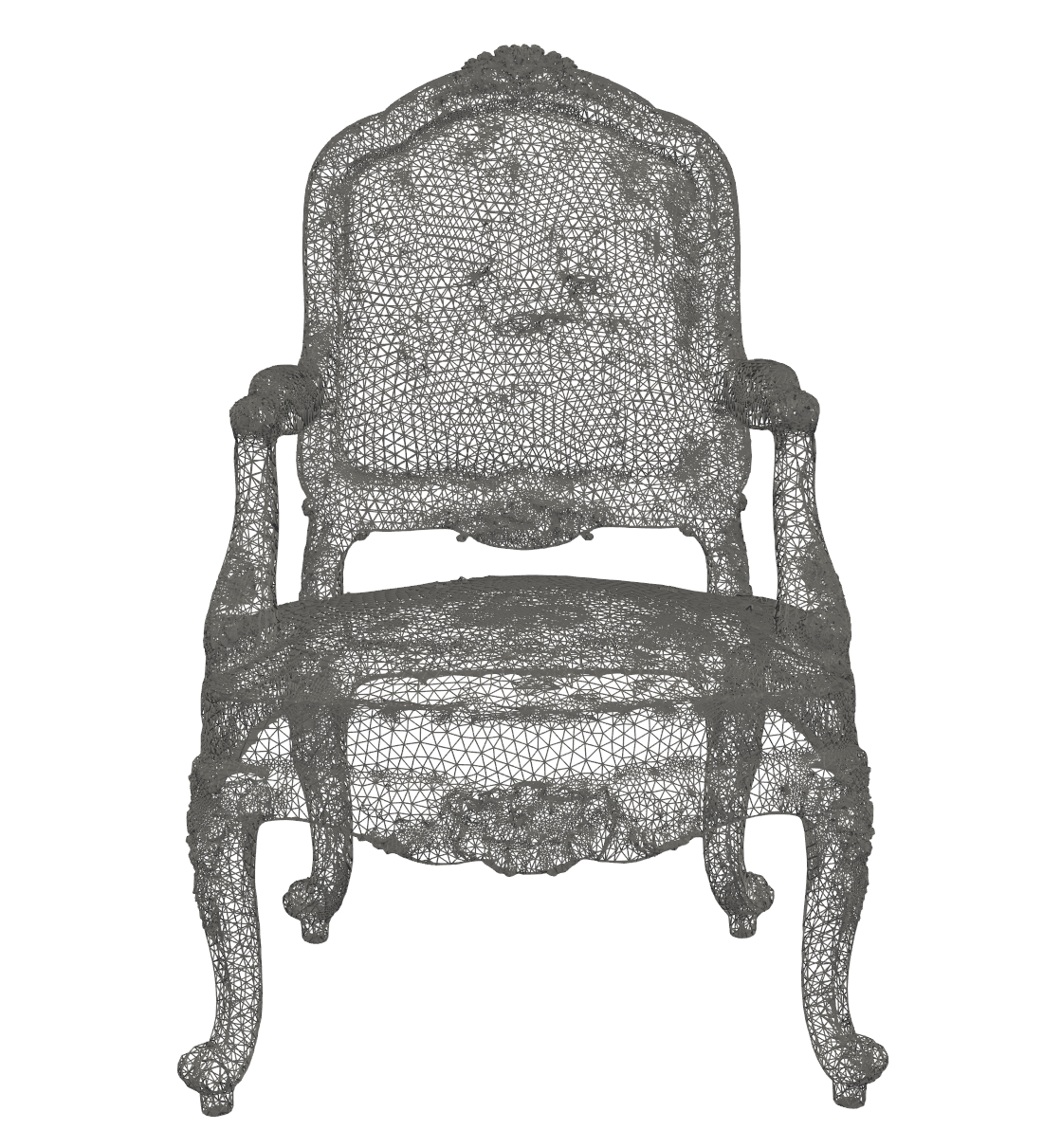} &
    \includegraphics[width=0.23\textwidth]{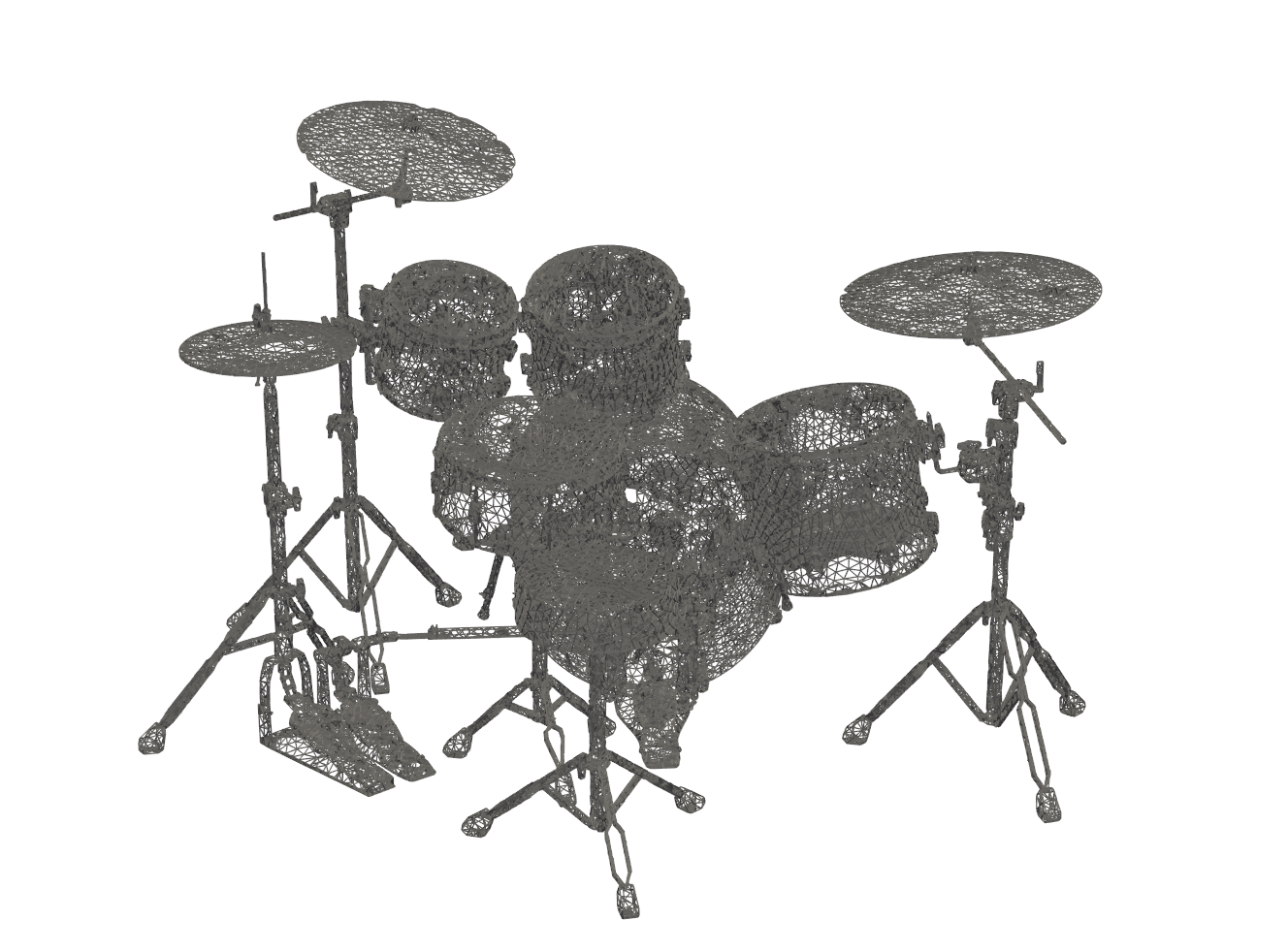} &
    \includegraphics[width=0.23\textwidth]{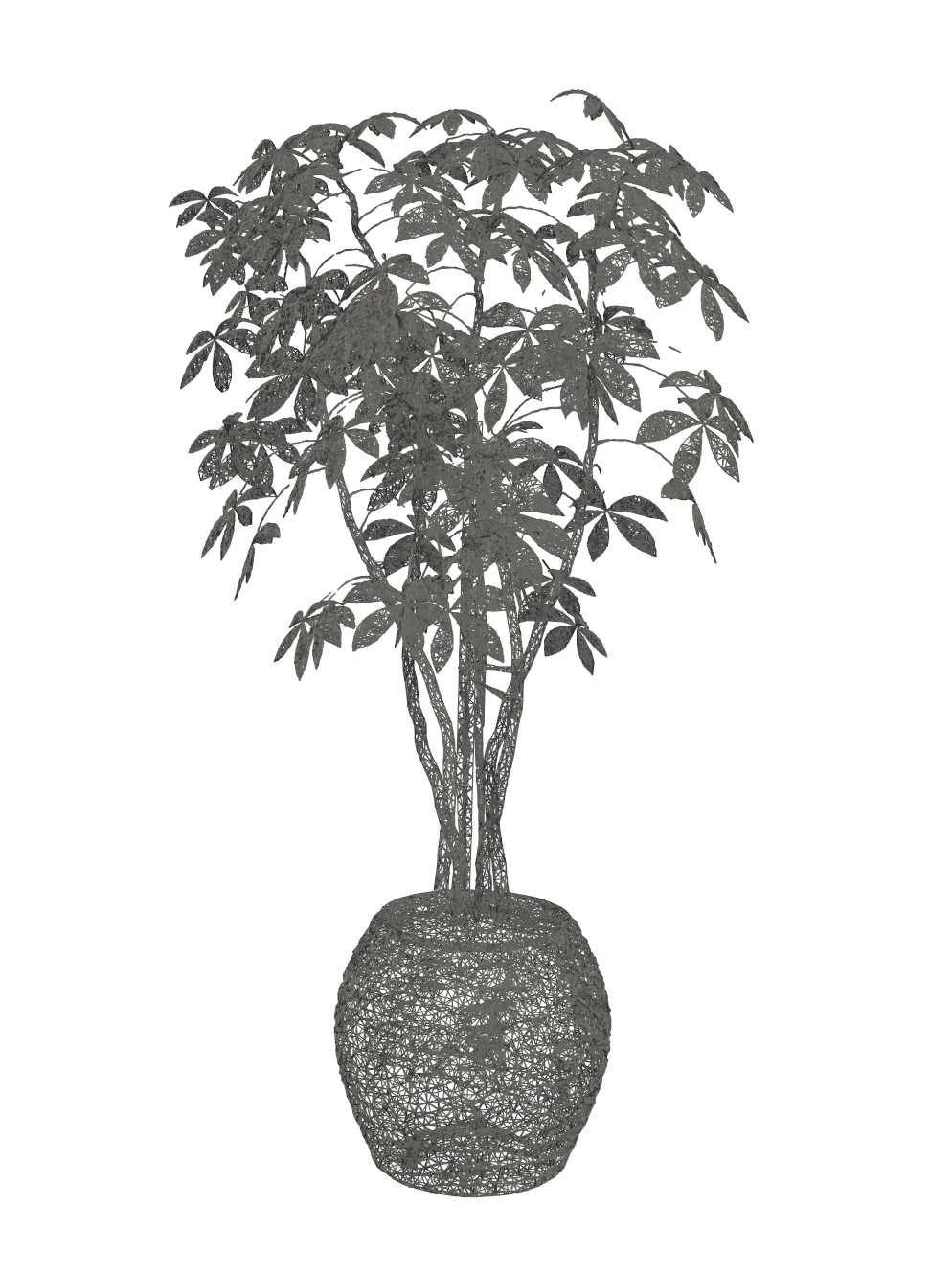} &
    \includegraphics[width=0.23\textwidth]{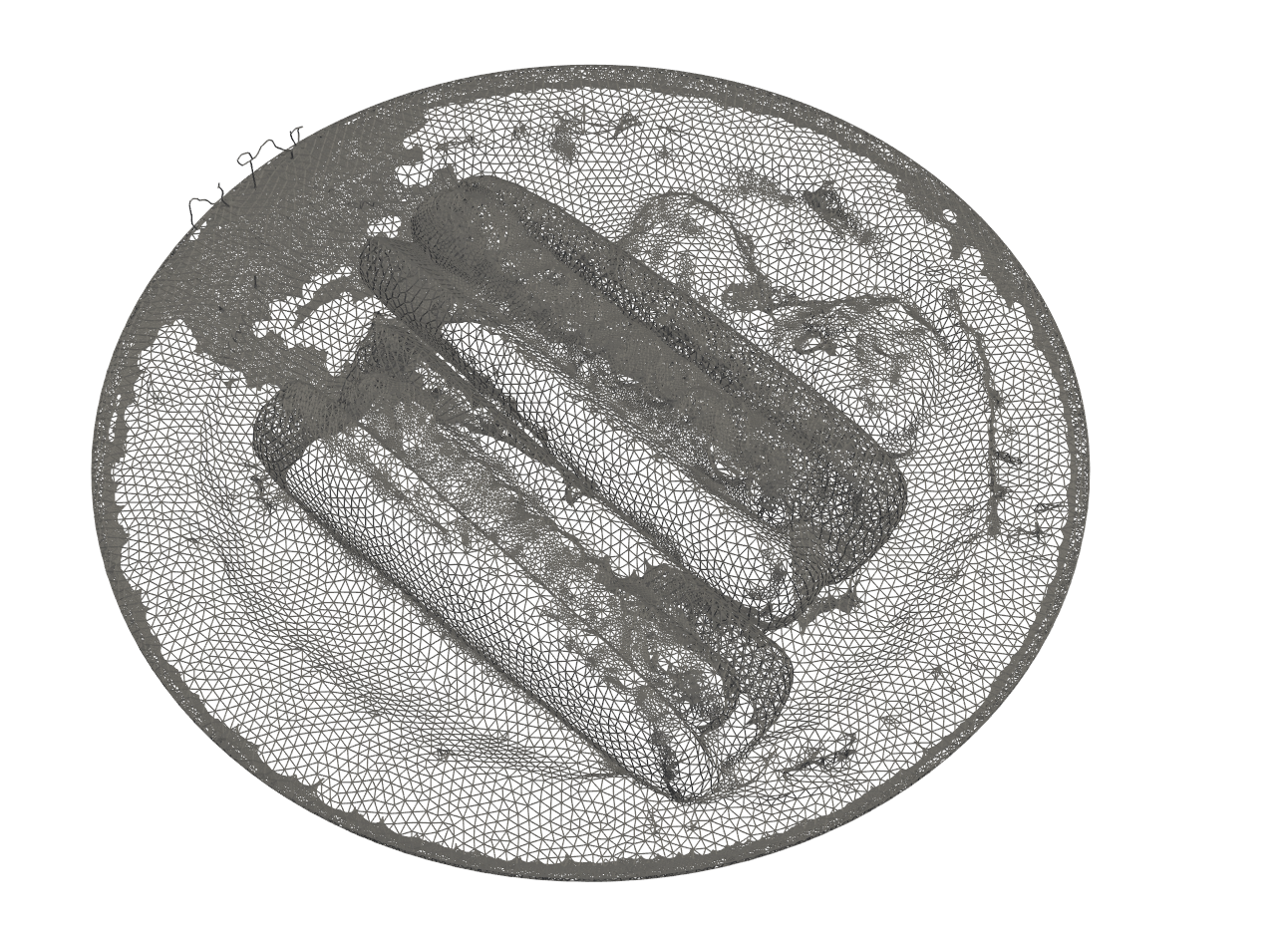} \\[-2pt]
    \includegraphics[width=0.23\textwidth]{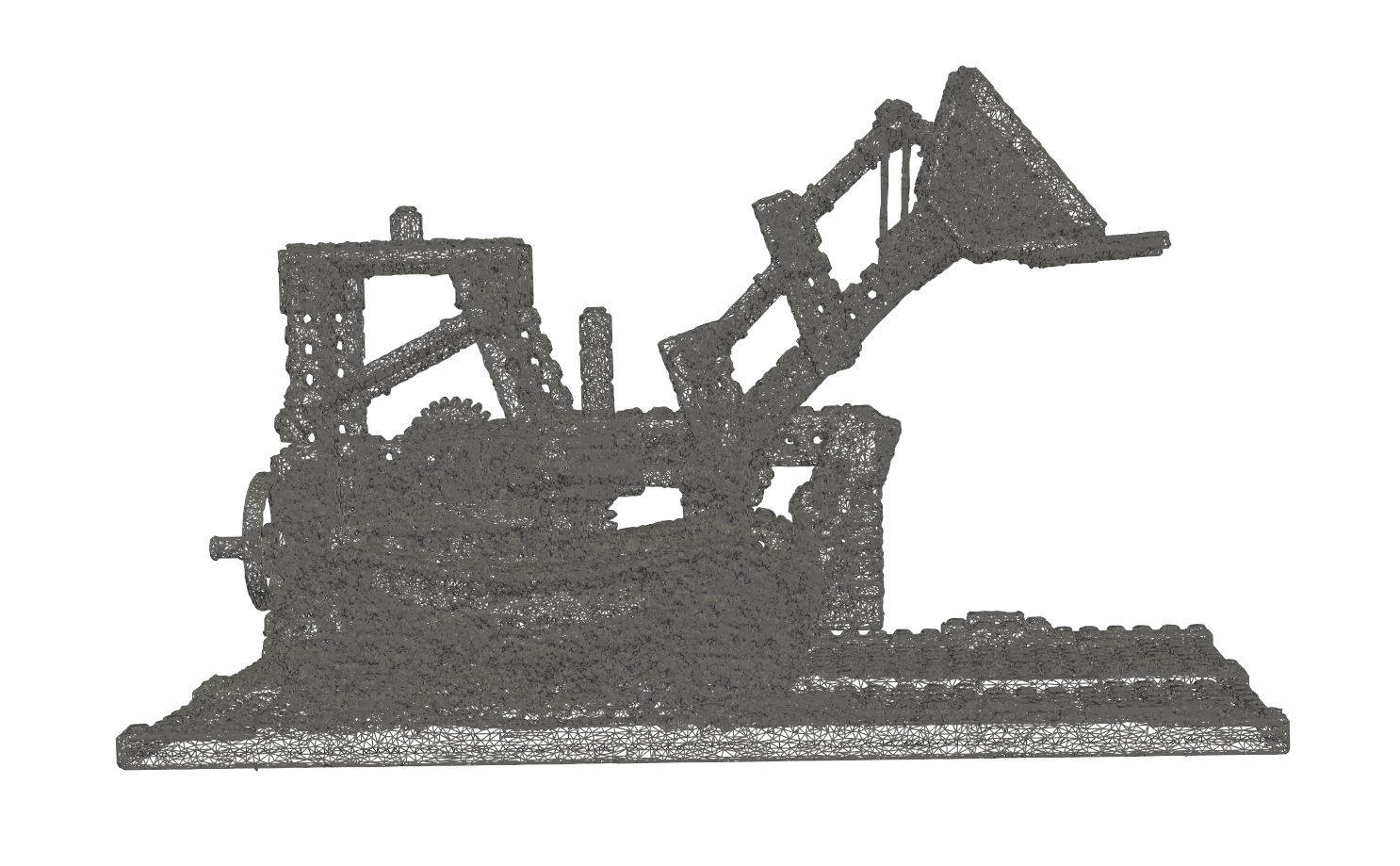} &
    \includegraphics[width=0.23\textwidth]{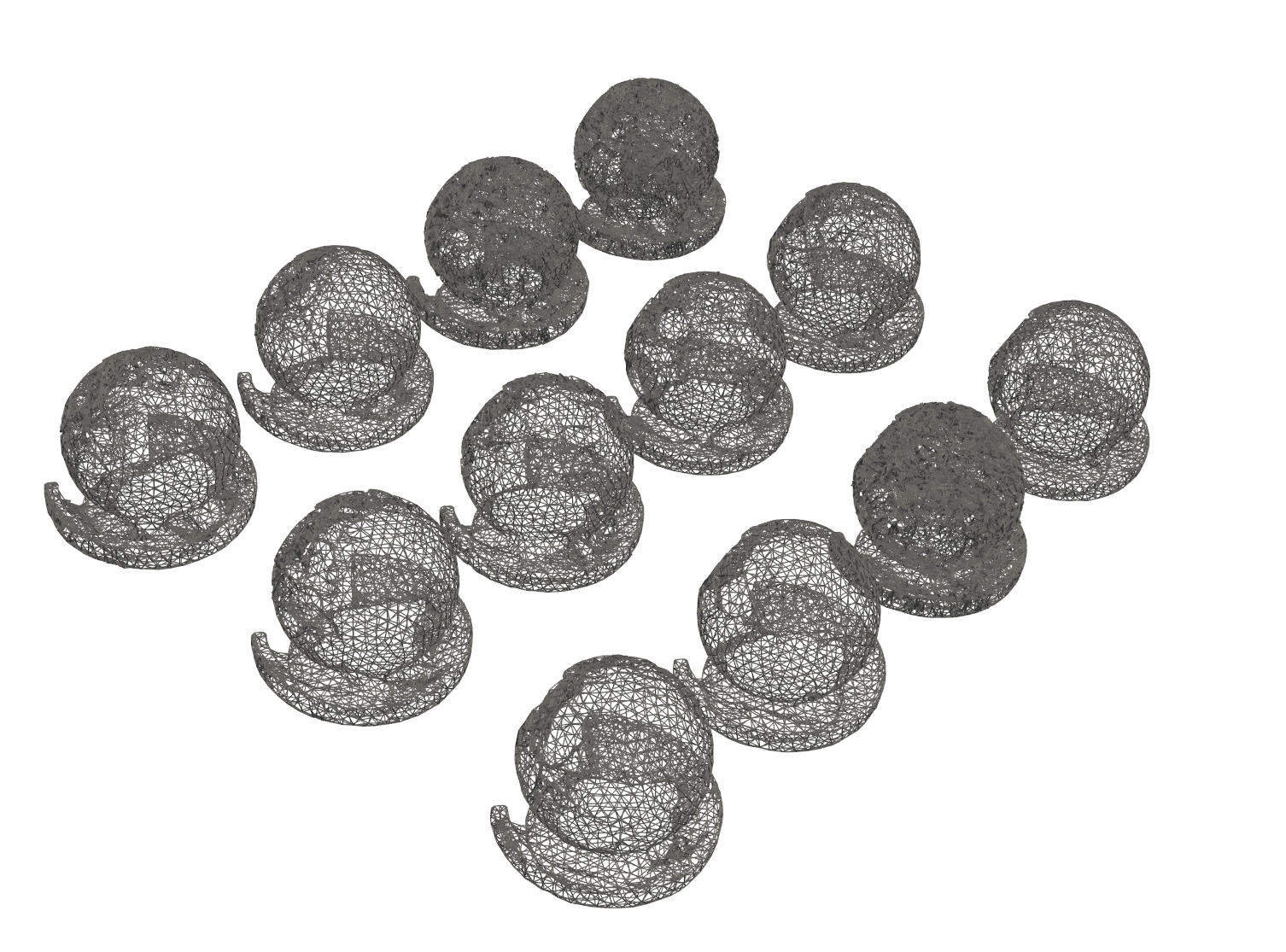} &
    \includegraphics[width=0.23\textwidth]{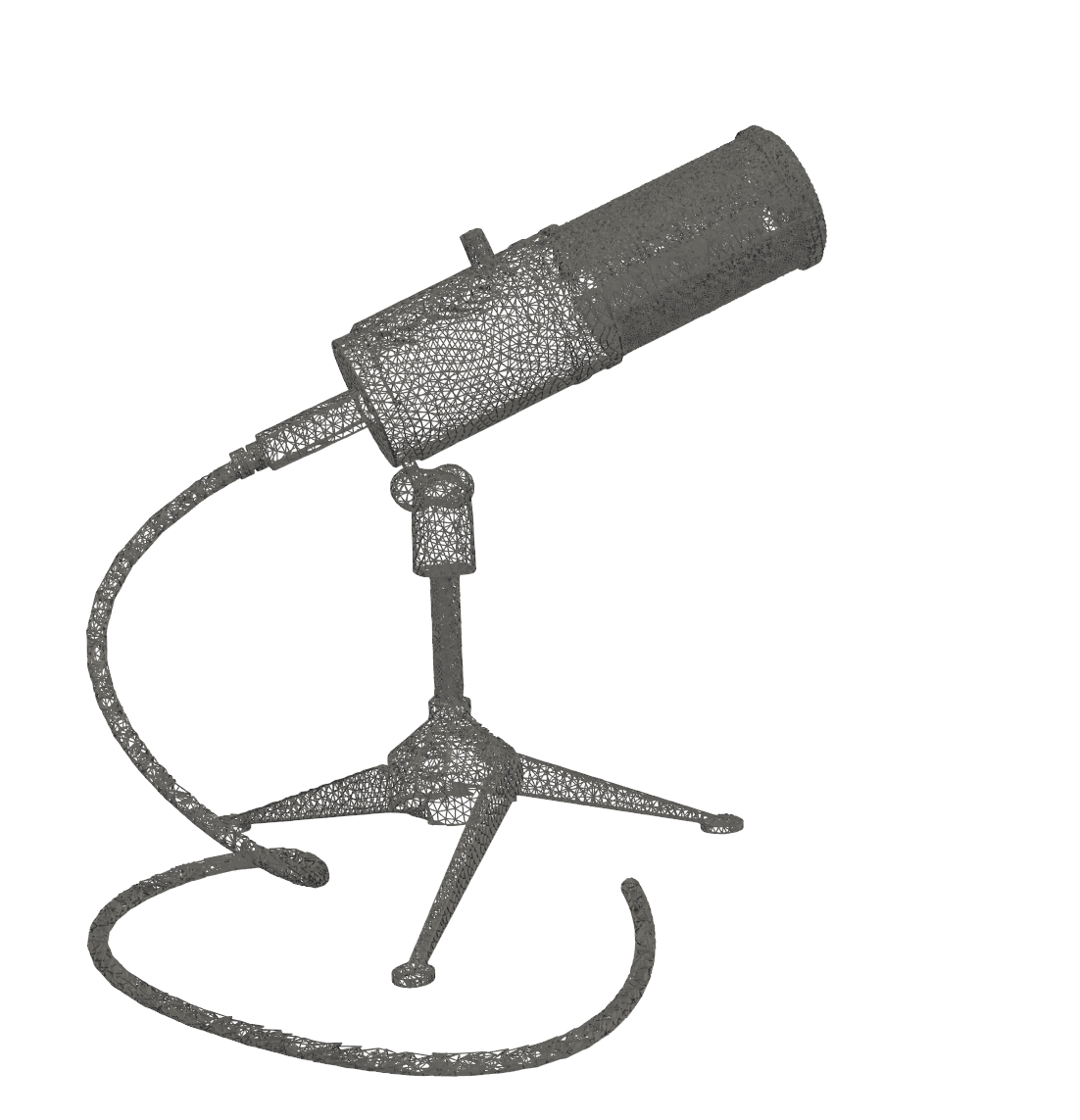} &
    \includegraphics[width=0.23\textwidth]{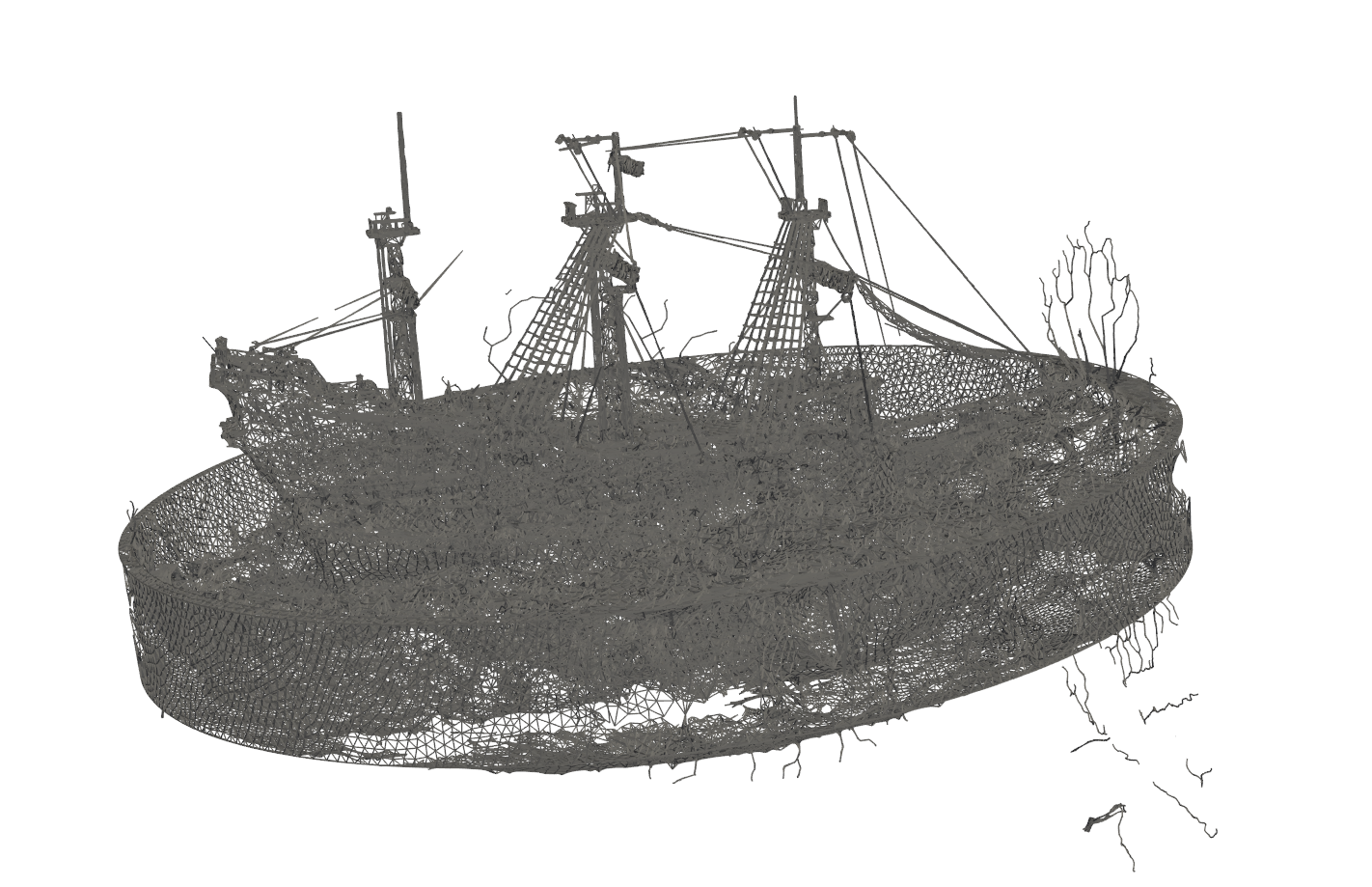} \\[3pt]
    \footnotesize Chair & \footnotesize Drums & \footnotesize Ficus & \footnotesize Hotdog \\
    \footnotesize Lego & \footnotesize Materials & \footnotesize Mic & \footnotesize Ship \\
  \end{tabular}
  \vspace{-4pt}
  \caption{Qualitative analysis of surface mesh reconstruction quality on the Blender (NeRF-Synthetic) dataset. 
  RePose-NeRF generates geometrically accurate and smooth meshes with refined structural consistency across multiple object categories.}
  \label{fig:surface_mesh_quality}
  \vspace{-4pt}
\end{figure*}
\FloatBarrier

\section{Limitations and Conclusion}
Although our method demonstrates promising results, it has certain limitations. While the pose refinement network can effectively correct moderately noisy camera poses, it assumes relatively smooth camera trajectories and struggles when pose noise exceeds 15°. Additionally, our approach shows reduced performance under extremely sparse view conditions. In future work, we aim to leverage data augmentation to generate synthetic views from the available sparse inputs, which could further improve reconstruction quality.

In summary, we propose an efficient framework for reconstructing textured surface meshes from multi-view RGB images with noisy camera poses. Our approach leverages NeRF to learn a neural implicit representation of the scene, employs hash-grid encoding from Instant-NGP for efficient positional encoding, and extracts detailed polygonal meshes with textured appearance. The resulting reconstructions exhibit high-quality geometry and textures, making them suitable for direct use in robotics and other downstream applications.

{
    \small
    \bibliographystyle{ieeenat_fullname}
    \bibliography{main}
}


\end{document}